\RequirePackage{fix-cm}

\documentclass[twocolumn,natbib]{svjour3}          
\smartqed  
\def\makeheadbox{\relax}
\usepackage{graphicx}
\usepackage{mathptmx}      
\usepackage{amsmath,amsfonts}
\usepackage{bbm}
\usepackage[T1]{fontenc}
\usepackage[latin9]{inputenc}
\usepackage{array}
\usepackage{float}
\usepackage{url}
\usepackage{multirow}
\usepackage{setspace}

\makeatletter
\usepackage{marvosym}
\newcommand{\envelope}{(\raisebox{-.5pt}{\scalebox{1.45}{\Letter}}\kern-.4pt )}
\begin{document}

\title{A Bayesian Method for Joint Clustering of Vectorial Data and Network Data
}


\author{Yunchuan Kong         \and
        Xiaodan Fan 
}

\authorrunning{Kong and Fan} 

\institute{Y. Kong \at
			Department of Biostatistics and Bioinformatics, Emory University, 1518 Clifton Rd, Atlanta, GA 30322, USA \\
              \email{yunchuan.kong@emory.edu}  
           \and
           X. Fan \envelope \at
              Department of Statistics, The Chinese University of Hong Kong, Shatin, New Territories, Hong Kong SAR, China \\
              \email{xfan@sta.cuhk.edu.hk}\\
}

\date{}

\maketitle

\begin{abstract}
We present a new model-based integrative method for clustering
objects given both vectorial data, which describes the feature of
each object, and network data, which indicates the similarity of
connected objects. The proposed general model is able to cluster
the two types of data simultaneously within one integrative
probabilistic model, while traditional methods can only handle one
data type or depend on transforming one data type to another.
Bayesian inference of the clustering is conducted based on a
Markov chain Monte Carlo algorithm. A special case of the general
model combining the Gaussian mixture model and the stochastic
block model is extensively studied. We used both synthetic data
and real data to evaluate this new method and compare it with
alternative methods. The results show that our simultaneous
clustering method performs much better. This improvement is due to
the power of the model-based probabilistic approach for
efficiently integrating information. 
\end{abstract}

\section{Introduction}
\label{intro} In social and economic life, as well as in many research fields such as data mining, image processing and bioinformatics, we often have the need to separate a set of objects to different groups according to their similarity to each other, so that we can subsequently represent or process different groups according to their different characteristics. As a unsupervised learning approach catering this general need, clustering data analysis has been extensively used in research and
real life \citep{jain2010data}.

In the Big Data era, complicated systems are often measured from
multiple angles. As a result, the same set of objects is often
described by both their individual characteristics and their
pairwise relationship. Often, the two types of data are from
different sources. For example, companies, such as Amazon and
Netflix, often need to divide customers into groups of different
consumption patterns, so that they can correctly recommend
commodities to a certain customer. In this scenario, the personal
information of a customer, such as the age and historical shopping
records, is the vectorial data that we can use for clustering. The
interrelationship between customers, such as how often two
customers shop together and how often they like same Facebook
posts, is the network data that can be used for clustering. For
another instance, in bioinformatics research, we often need to
cluster genes into different groups, which ideally correspond to
different gene regulatory modules or biochemical functions. In
this scenario, the expression of genes under different conditions, such
as microarray data from different tissues or different
environmental stimulus, is the vectorial data for gene clustering.
The network data for gene clustering includes gene regulatory networks, protein-protein interaction data and whether a pair of genes belongs to a same Gene Ontology group. Therefore, we have to integrate both vectorial data and network data to better elucidate
the group structure among the objects.

Traditional clustering methods are designed for either vectorial
data alone or network data alone \citep{buhmann1995data}. In the
most commonly used vectorial data, each object is represented by a
vector of the same dimension. The similarity between two objects
is reflected by certain distance measure of the two corresponding
vectors. The problem of clustering vectorial data have been
studied for more than 60 years. The most widely used methods
include K-means clustering
\citep{macqueen1967some,tavazoie1999systematic}, Gaussian mixture
model \citep{mclachlan1988mixture,fraley2002model} and
hierarchical clustering
\citep{sibson1973slink,defays1977efficient,eisen1998cluster}. Most
vectorial data clustering methods adopted a central clustering
approach by searching for a set of prototype vectors.
\citet{jain2010data} provided a good review on this subject. The
other data type is network data, where the similarity between two
objects is directly given without describing the characteristics
of individual objects. The problem of clustering network data
arose only in the recent two decades. The methods for network data
clustering include entropy-based methods
\citep{buhmann1994maximum,park2006validation}, spectrum-based
methods \citep{swope2004describing,bowman2009using}, cut-based
methods \citep{muff2009identification}, path-based method
\citep{jain2012identifying}, modified self-organizing maps
\citep{seo2004self}, mean field model \citep{hofmann1997pairwise},
probabilistic relational model \citep{taskar2001probabilistic,
nowicki2001estimation, mariadassou2010uncovering}, and Newman's
modularity function \citep{newman2006modularity}.
\citet{fortunato2010community} provided a good review on this
subject.

As shown above, vectorial data clustering and network data
clustering have both been intensively studied. In contrast, as to
our knowledge, no existing methods can integrate the clustering
information in the two individual data types parallelly within a
coherent framework. There are several papers on the direction of
data integration which can take both vectorial and network data as
input, but these methods either transform the network data to
vectorial data as in the latent position space approach
\citep{hoff2002latent, handcock2007model, gormley2011mixture}, or
transform the vectorial data to network data as in
\citet{zhou2010clustering} and \citet{gunnemann2010subspace,
gunnemann2011db}. The explicit or implicit data transformation
needs an artificial design of a latent metric space for converting
the network data or an artificial design of a distance measure for
converting the vector data. Thus they cannot avoid the arbitrary
weighting of the clustering information from two data types. In
reality, we seldom know how to weight one data type again another.
For example, the vectorial data and the network data may come from
independent studies which may have used different techniques to
check the similarity of objects at different levels. Thus,
essentially we have no good way to weight one against the other.
The only common and comparable thing behind the two data types is
how likely each pair of objects is within a same cluster.

In this paper, we developed an integrative probabilistic
clustering method called ``Shared Clustering'' for clustering
vectorial data and network data simultaneously. We assume that the
vectorial data is independent from the network data conditional on
the cluster labels. Our probabilistic model treats the two types
of data equally instead of treating one as the covariate of the
other, and models their contribution to clustering directly
instead of converting one type to another. We perform the
statistical inference in the Bayesian framework. The Markov chain
Monte Carlo (MCMC) algorithm, or more specifically the Gibbs
sampler, is employed to sample the parameters and cluster labels.
The paper is organized as follows. We first describe the model of
Shared Clustering, then the inference method is described in
detail, followed by applications to both synthesized data and real
data. A summary and discussion are provided at the end.

\section{Problem statement and model specification}
\label{probl} We consider the clustering of $N$ objects according
to their vectorial data $\mathbf{x}_i$ and pairwise data
$\mathbf{y}_{ij}$, where\emph{ }$i,j=1,\ldots, N$ are the indexes
of objects. Let $\boldsymbol X$ be the $N$\emph{-}by-$q$\emph{
}matrix formed by $\mathbf{x}_i$ whose dimension is $q$, and
$\boldsymbol Y$ be the $N$\emph{-}by-$N$\emph{ }square matrix
formed by $\mathbf{y}_{ij}$. Note that $\mathbf{y}_{ij}$ can be
deemed as the weight of the link from the $i$-th object to the
$j$-th object on the network. In the Shared Clustering model, we
assume that the vectorial data $\boldsymbol X$ and the network
data $\boldsymbol Y$ share a common clustering structure
$\boldsymbol{C}=(c_{1},\ldots,c_{N})$, where the cluster label of
the $i$-th object is $c_{i}=1,\ldots, K$, and $K$ is the total
number of clusters. Given $\boldsymbol C$, all $\mathbf{x}_i$ and
$\mathbf{y}_{ij}$ are assumed to be independently following their
corresponding component distributions. Thus, the joint likelihood
function is $L(\boldsymbol X,\boldsymbol Y|\boldsymbol \Phi,
\boldsymbol \Psi, \boldsymbol C) = \prod\limits_{i=1}^N
f(\mathbf{x}_{i}|\phi_{c_i}) \cdot \prod\limits_{i=1}^N
\prod\limits_{j=1}^N g(\mathbf{y}_{ij}|\psi_{c_i,c_j})$, where
$\boldsymbol\Phi=(\phi_{1},\ldots,\phi_{K})$ and
$\boldsymbol\Psi=(\psi_{1},\ldots,\psi_{K})$ represent all
component specific parameters, $f(\cdot)$ and $g(\cdot)$ represent
the component distributions.

We further assume that each of the $N$ cluster labels follows a
multinomial distribution with the probability vector $\boldsymbol
P=(p_{1},\ldots,p_{K})\in\mathbbm{R}^K$, namely $c_i=k$ with
probability $p_k$, $k=1,\ldots,K$. Intuitively, the meaning of
$\boldsymbol P$ is the prior probabilities that each object is
assigned to the corresponding clusters.

In summary, the generative version of the model can be stated
as:
\begin{equation}\begin{split}&
c_i\sim\mathrm{Multinomial}(\boldsymbol P),\\&
\mathbf{x}_i|c_{i}\sim\mathrm{f}(\mathbf{x}_{i}|\phi_{c_i}),\\&
\mathbf{y}_{ij}|c_{i},c_j\sim\mathrm{g}(\mathbf{y}_{ij}|\psi_{c_i,c_j}).
\end{split}\label{generative}\end{equation}

\begin{figure}[h]
\noindent \begin{centering}
\includegraphics[scale=0.3]{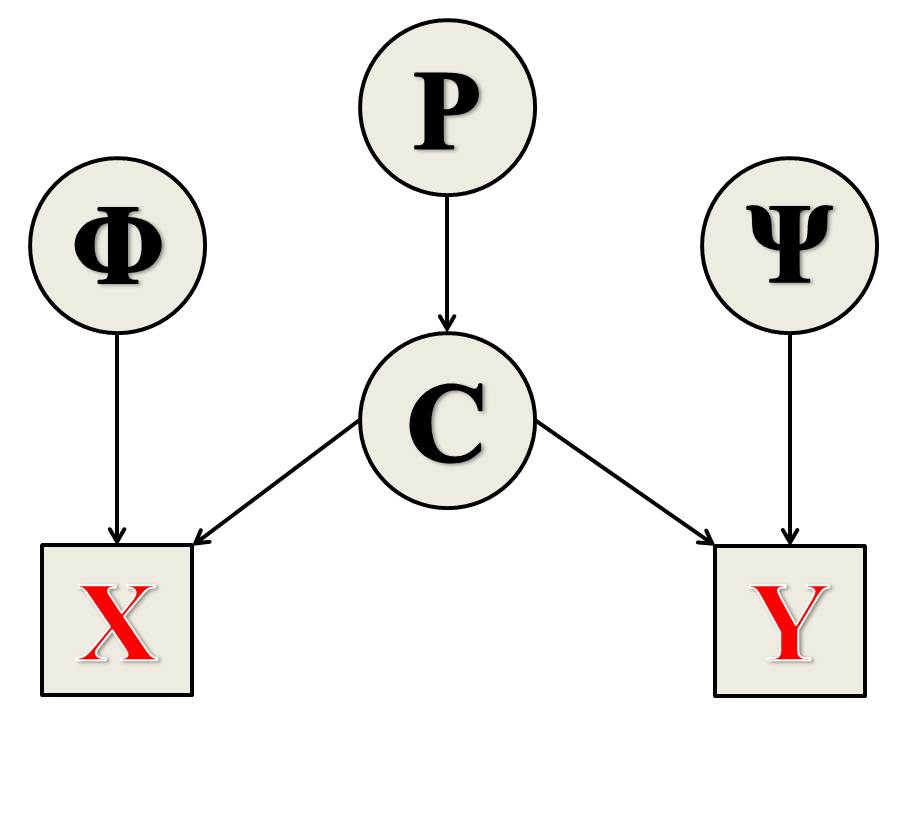}
\par\end{centering}
\caption{Dependency structure of the variables in Shared Clustering}
\label{depstruc}
\end{figure}

The dependency structure of all random variables is shown in Fig.
\ref{depstruc}.

The general joint clustering model in Equation \eqref{depstruc}
conveys the main idea to integrate the model for vectorial data
and the model for network data probabilistically by conditioning
on shared cluster labels, no matter what models are used for
individual data types. The component distributions of $\boldsymbol
X$ and $\boldsymbol Y$ can be any distribution combinations
depending on the specific types of the given data. For example,
$f(\cdot)$ can be either a continuous distribution or a discrete
distribution depending on the given vectorial data. A proper
distribution $g(\cdot)$, say Poisson distribution, may be induced
to model network variables if an integer weighted graph is given.

For a concrete study of the joint clustering model, we assume that
vectorial data follows a Gaussian Mixture Model (GMM)
\citep{mclachlan1988mixture,fraley2002model} and the network data
follows a Stochastic Block Model (SBM)
\citep{nowicki2001estimation}. More specifically, we assume that
the vectorial data $\mathbf{x}_{i}$ follows a multivariate Normal
distribution with $\phi_{c_i}=(\mu_{c_i},\Sigma_{c_i})$, and
$\mathbf{y}_{ij}$ is a binary variable following Bernoulli
distribution with linking probability equal to $\psi_{c_i,c_j}$.
Here the network interested is an undirected graph without self
loop, thus $\boldsymbol Y$ is an $N$-by-$N$ symmetric matrix with
all diagonal entries being zero. In this remaining part of this
paper, we will mainly focus on these two specified distribution
assumptions for $\boldsymbol X$ and $\boldsymbol Y$ respectively,
and we call this combination the Normal-Bernoulli model.

Under the Normal-Bernoulli model, the conditional distribution of
each vector $\mathbf{x}_i\in\mathbbm{R}^q,i=1,\ldots,N$ given
$c_{i}$ is
\begin{equation}\mathbf{x}_i|c_{i}\sim\mathrm{N}(\mu_{c_{i}},\Sigma_{c_{i}}),\label{x_i|c_i}\end{equation}
where $\mu_{c_{i}}\in\mathbbm{R}^q$ is the mean vector and
$\Sigma_{c_{i}}$ is the $q$-by-$q$ covariance matrix. From
Equation \eqref{x_i|c_i}, $\mathbf{x}_i$ belongs to the $k$-th
cluster if and only if $c_{i}=k$, where $k=1,\ldots,K$.

In SBM, a network is partitioned into several blocks according to
the number of total clusters $K$. Variables $y_{ij}$ within each
individual block are controlled by a same set of parameters. In
our case the parameter $\boldsymbol \Psi$ for network data is
therefore a $K$-by-$K$ probability matrix, with each element
describing the corresponding Bernoulli distribution within a
certain block. Thus the distribution of the edge variable
$y_{ij},i,j=1,\ldots,N,i\neq j$ is
\begin{equation}y_{ij}|c_i,c_j\sim
\mathrm{Bernoulli}(\psi_{c_i,c_j}),\label{y_{ij}|c_i,c_j}\end{equation}
namely $g(y_{ij}|\boldsymbol
\Psi,c_{i},c_{j})=\psi_{c_i,c_j}^{y_{ij}}\cdot(1-\psi_{c_i,c_j})^{1-y_{ij}}$.
In terms of undirected networks, the network data $\boldsymbol Y$
or the adjacency matrix of the graph is symmetric. Hence we have
$y_{ij}=y_{ji}$ and $\psi_{c_i,c_j}=\psi_{c_j,c_i}$. Thus only the
lower (or upper) triangles of $\boldsymbol Y$ and
$\boldsymbol\Psi$ need to be considered.

Given a dataset $\boldsymbol D=(\boldsymbol X,\boldsymbol Y)$ of $N$ objects defined as above and the total number of clusters $K$, our task is to infer the true cluster membership $\boldsymbol C$. In other words, we work on how the $N$ objects should be divided
into $K$ clusters according to the integrative information of their vectorial data and network data.

\section{Method description}
\label{metho}
\subsection{Prior distributions}
\label{prior} For Bayesian inference, we need to specify prior
distributions for unknown parameters. In the case that little
prior knowledge about the parameters is available, we choose flat
priors as in most Bayesian data analyses. Meanwhile, we would like
to use fully conjugate priors to ease the posterior sampling.
Although different prior sets could be assigned to the $K$
different cluster components, we use the same prior settings in
absence of the prior knowledge of the $K$ different component
distributions.

As stated in Section \ref{probl}, the cluster labels in
$\boldsymbol C$ follow a multinomial distribution. One common way
is to fix $p_i=1/K$, but this indicates a strong prior belief that
each cluster is of equal size. Thus we instead treat all $p_i$ as
unknown and assume the vector $\boldsymbol P=(p_{1},\ldots,p_{K})$
follows a Dirichlet distribution with prior parameter vector
$\boldsymbol a\in\mathbbm{R}^K$, i.e., $\boldsymbol
P\sim\mathrm{Dirichlet}(\boldsymbol a)$.

As for the multivariate Normal distributions, a conventional fully conjugate prior setting discussed in \citet{Rossi2006GMM}, which is a special case of multivariate regression, is to assume that the mean vector $\mu_k$ follows multivariate Normal distribution given the covariance matrix $\Sigma_k$, and $\Sigma_k$ follows Inverse-Wishart distribution. Namely, $\Sigma_k\sim\mathrm{IW_q}(T,v_0)$ and $\mu_k\sim\mathrm{N}(\mu_{0},\alpha^{-1}\Sigma_{k})$, where $T$ is the $q\times q$ location matrix of Inverse-Wishart prior on $\Sigma_k$, $v_0$ is the corresponding degree of freedom, $\mu_0$ is the mean of the multivariate Normal prior on $\mu_k$, and $\alpha$ is a precision parameter. In our experiments demonstrated in later sections, we will use the default priors described in \citet{Rossi2006GMM} and its R implementation \citep{bayesmR} for the vectorial data.

For the network data $\boldsymbol Y$, the conjugate prior for
individual Bernoulli parameter $\psi_{c_i,c_j}$ is Beta
distribution with shape parameters $\beta_1$ and $\beta_2$, i.e.,
$\psi_{c_i,c_j}\sim\mathrm{Beta}(\beta_1,\beta_2)$. And again we
uniformly set the same pair of $(\beta_1,\beta_2)$ for every
$\psi_{c_i,c_j}$ for the lack of nonexchangeable prior knowledge.
In our simulation studies, $\beta_1$ and $\beta_2$ are set both
equal to a relatively small quantity which is slightly larger than
1. Sensitivity analysis in Section \ref{sensi} shows the two prior
settings result in little disparity.

\subsection{Posterior distributions}
\label{poste} The full joint posterior distribution of all
parameters is proportional to the product of the joint likelihood
and the joint prior distributions, thus we have
\begin{equation}\begin{split}p(&\boldsymbol P,\boldsymbol
C,\boldsymbol \Phi,\boldsymbol \Psi|\boldsymbol X,\boldsymbol
Y)\\&\propto p(\boldsymbol X,\boldsymbol Y|\boldsymbol
P,\boldsymbol C,\boldsymbol \Phi,\boldsymbol \Psi)p(\boldsymbol
P,\boldsymbol C,\boldsymbol \Phi,\boldsymbol
\Psi)\\&=p(\boldsymbol X|\boldsymbol C,\boldsymbol
\Phi)p(\boldsymbol Y|\boldsymbol C,\boldsymbol \Psi)p(\boldsymbol
\Phi)p(\boldsymbol \Psi)p(\boldsymbol C|\boldsymbol
P)p(\boldsymbol P).\end{split}\label{full_post}\end{equation}

\subsection{Gibbs sampling algorithm}
\label{gibbs} We use Gibbs sampler to conduct the Bayesian
inference, which samples the parameters from their conditional
posterior distributions iteratively \citep{robert2004monte}. The
specified conditional posteriors of the model parameters for Gibbs
sampling are provided in the appendix.

Our algorithm is similar to the case of Gibbs sampling for GMM or
SBM alone, with the essential difference that the distributions of
cluster labels are now associated with the two types of data
jointly. A pseudo code of the algorithm is presented in Table
\ref{pseudo code}. 

\begin{center}
\begin{table*}
\begin{singlespace}
\noindent \centering{}%
\begin{tabular}{ll}
 & \textbf{Gibbs Sampler for Shared Clustering Model}
\tabularnewline 1: & Set all hyper-priors
$\mu_0$,$\alpha$,$T$,$v_0$,$\boldsymbol a$,$\beta_1$,$\beta_2$ and
initialize $\boldsymbol P$,$\boldsymbol C$\tabularnewline 2: &
\textbf{for} each iteration \textbf{do} \tabularnewline 3: & \quad
\textbf{for }$k=1,\ldots,K$\textbf{do} \tabularnewline 4: &
\quad\quad Under current cluster label $\boldsymbol C$, extract
the $k$-th component of $\boldsymbol X$ \tabularnewline 5: &
\quad\quad Sample $\Sigma_k$ using Equation \eqref{app_sigma_post}
\tabularnewline 6: & \quad\quad Sample $\mu_k$ using Equation
\eqref{app_mu_post} \tabularnewline 7: & \quad\quad
\textbf{for}$j=1,\ldots,K$ \textbf{do} \tabularnewline 8: &
\quad\quad\quad Under current cluster label $\boldsymbol C$,
extract the block of cluster $k$ and $j$ from $\boldsymbol Y$
\tabularnewline 9: & \quad\quad\quad Sample $\psi_{k,j}$ using
Equation \eqref{app_psi_post} \tabularnewline 10: & \quad\quad
\textbf{end for} \tabularnewline 11: & \quad\textbf{end for}
\tabularnewline 12: & \quad\textbf{for }$i=1,\ldots,N$\textbf{do}
\tabularnewline 13: & \quad\quad \textbf{for }$k=1,\ldots,K$
\textbf{do} \tabularnewline 14: & \quad\quad\quad Set $c_i=k$ and
calculate $p(\mathbf x_i,\mathbf y_i|\boldsymbol
\Phi,\boldsymbol\Psi,c_i=k,C_{-i})p(c_i=k|\boldsymbol
P)$\tabularnewline 15: & \quad\quad \textbf{end for}
\tabularnewline 16: & \quad\quad Sample $c_i$ from $p(\mathbf
x_i,\mathbf y_i|\boldsymbol
\Phi,\boldsymbol\Psi,c_i,C_{-i})p(c_i|\boldsymbol P)$ after
normalizing\tabularnewline 17: & \quad \textbf{end
for}\tabularnewline 18: & \quad Sample $\boldsymbol P$ using
Equation \eqref{app_p_post} \tabularnewline 19: & \quad Calculate
the unnormalized joint posterior probability as in Equation
\eqref{full_post} \tabularnewline 20: & \textbf{end
for}\tabularnewline
\end{tabular}
\caption{Pseudo code of the algorithm}
\label{pseudo code}
\end{singlespace}
\end{table*}
\par\end{center}

After running the chain until convergence, the remaining
iterations after burn-in are used for posterior inference. More
specifically, when a point estimate of the clustering label
$\boldsymbol C$ is needed, we use the maximum a posteriori (MAP)
estimation, i.e. the iteration with the maximal joint posterior
probability \citep{sorenson1980parameter}. Using MAP can bypass
the label-switching problem. To quantify the clustering
uncertainty, we use the whole converged sample by summarizing it
in a heatmap of the posterior pairwise co-clustering probability
matrix. This heatmap can provide us a way of selecting the number
of clusters $K$ (see Section \ref{selec}).

\section{Synthetic data experiments}
\label{simul}
\subsection{Experimental design}
\label{exper}
We test the performance of our method under diverse scenarios. The difficulty of a clustering problem is determined by many factors, including the number of clusters $K$, the number of objects $N$, the tightness of clusters and the relative locations of clusters. We design different difficulty levels for $\boldsymbol X$ and $\boldsymbol Y$ separately, and test on their combinations.

For the vectorial data $\boldsymbol X$, we tried three different shapes (denoted as shape=1,2,3) and two overlapping conditions (with or without overlap), which are shown in Fig. \ref{type and ovl}. Corresponding parameters are listed in Table \ref{para for type and ovl}. For easy visualization, these examples are limited as two-dimensional. Higher dimensional cases are tested in Section \ref{simu3}.

\begin{center}
\begin{figure*}
\begin{centering}
\includegraphics[scale=0.3]{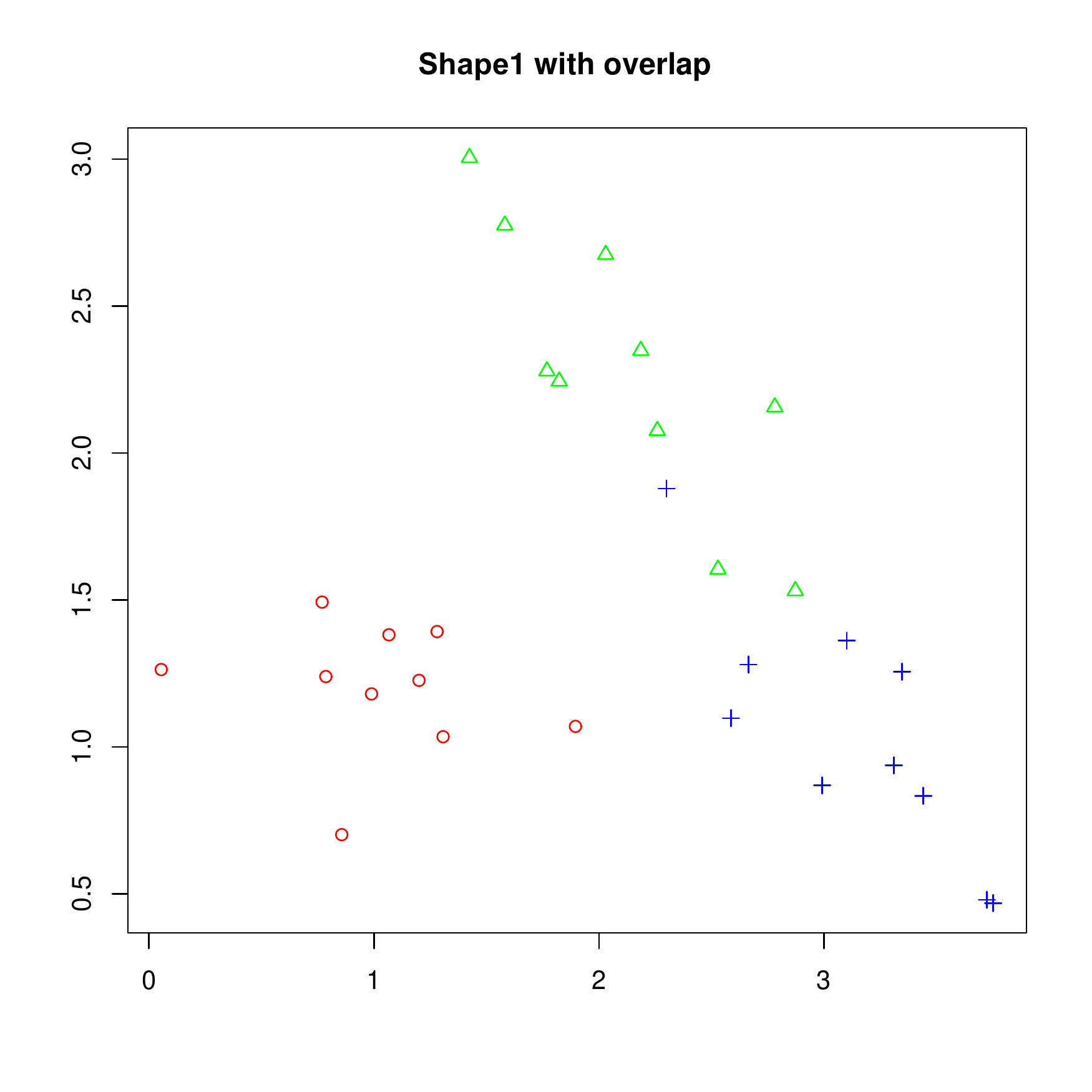}
\includegraphics[scale=0.3]{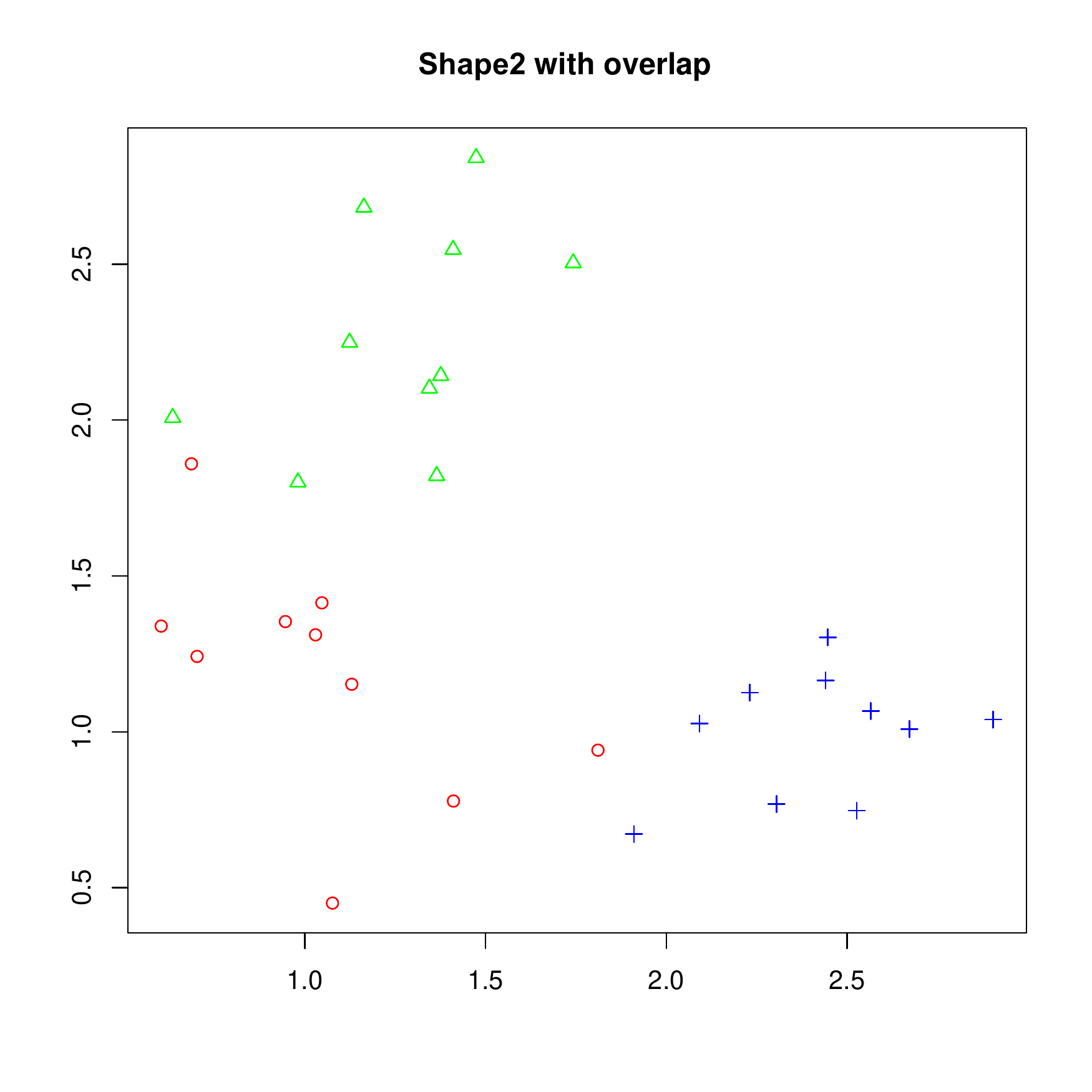}
\includegraphics[scale=0.3]{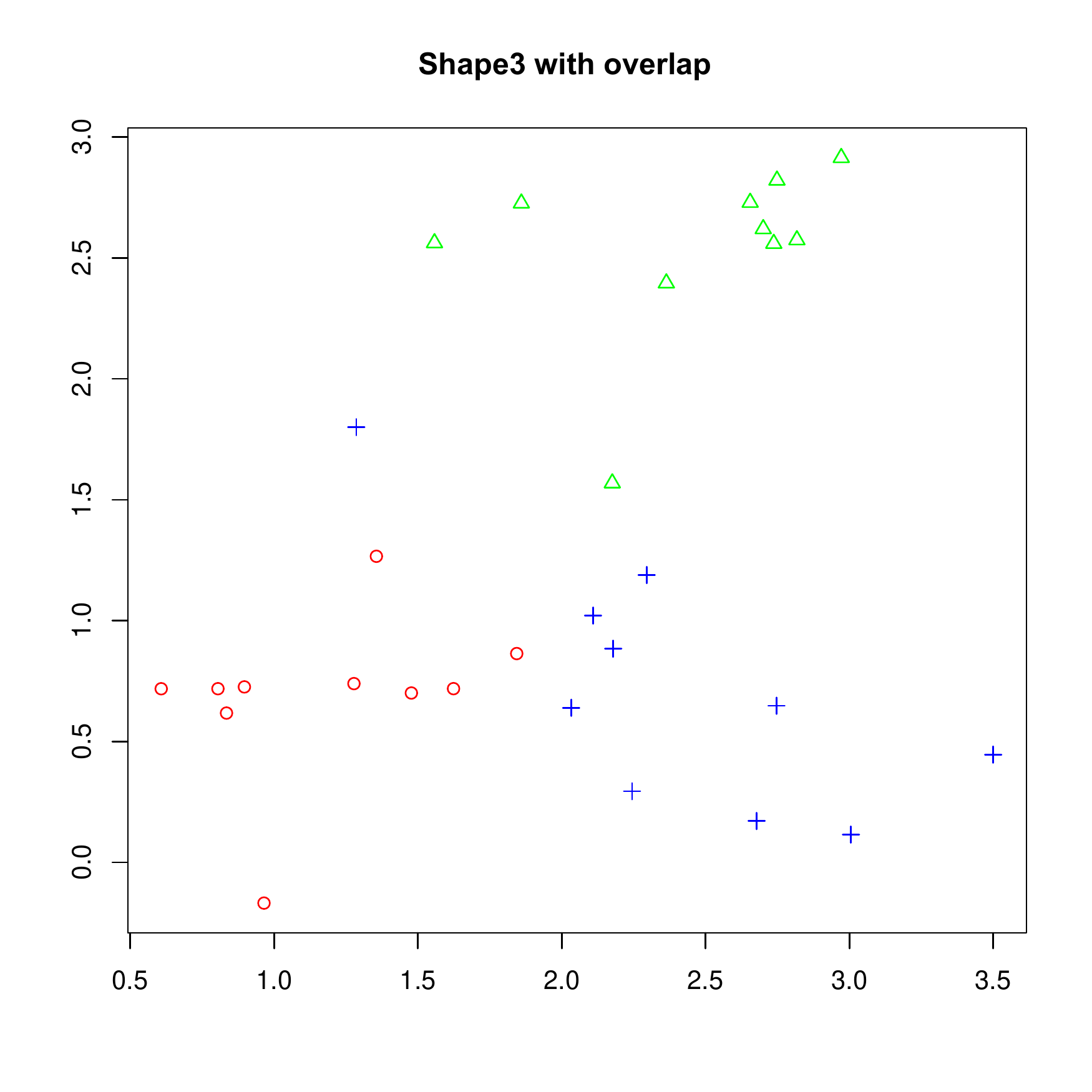}
\par\end{centering}
\begin{centering}
\includegraphics[scale=0.3]{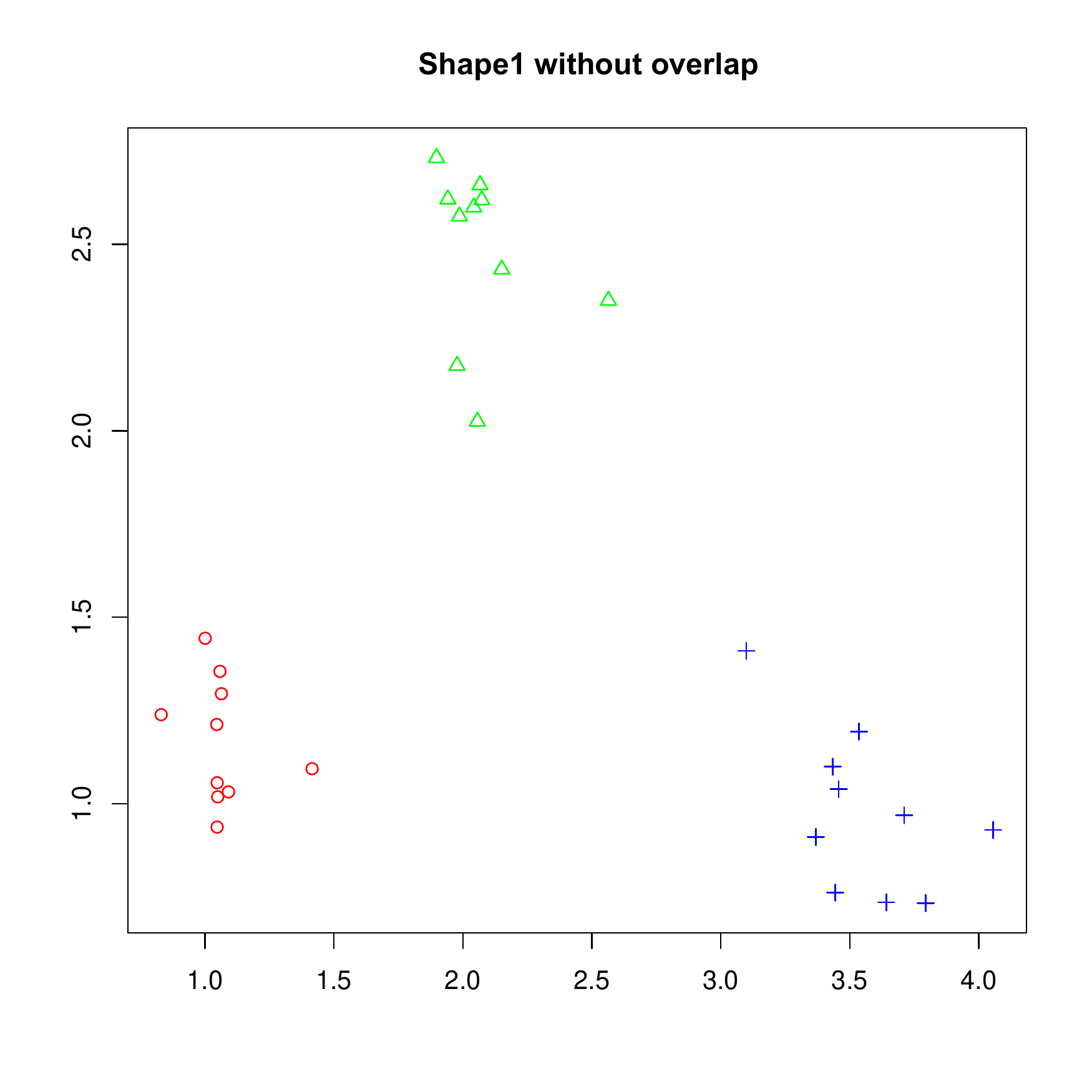}
\includegraphics[scale=0.3]{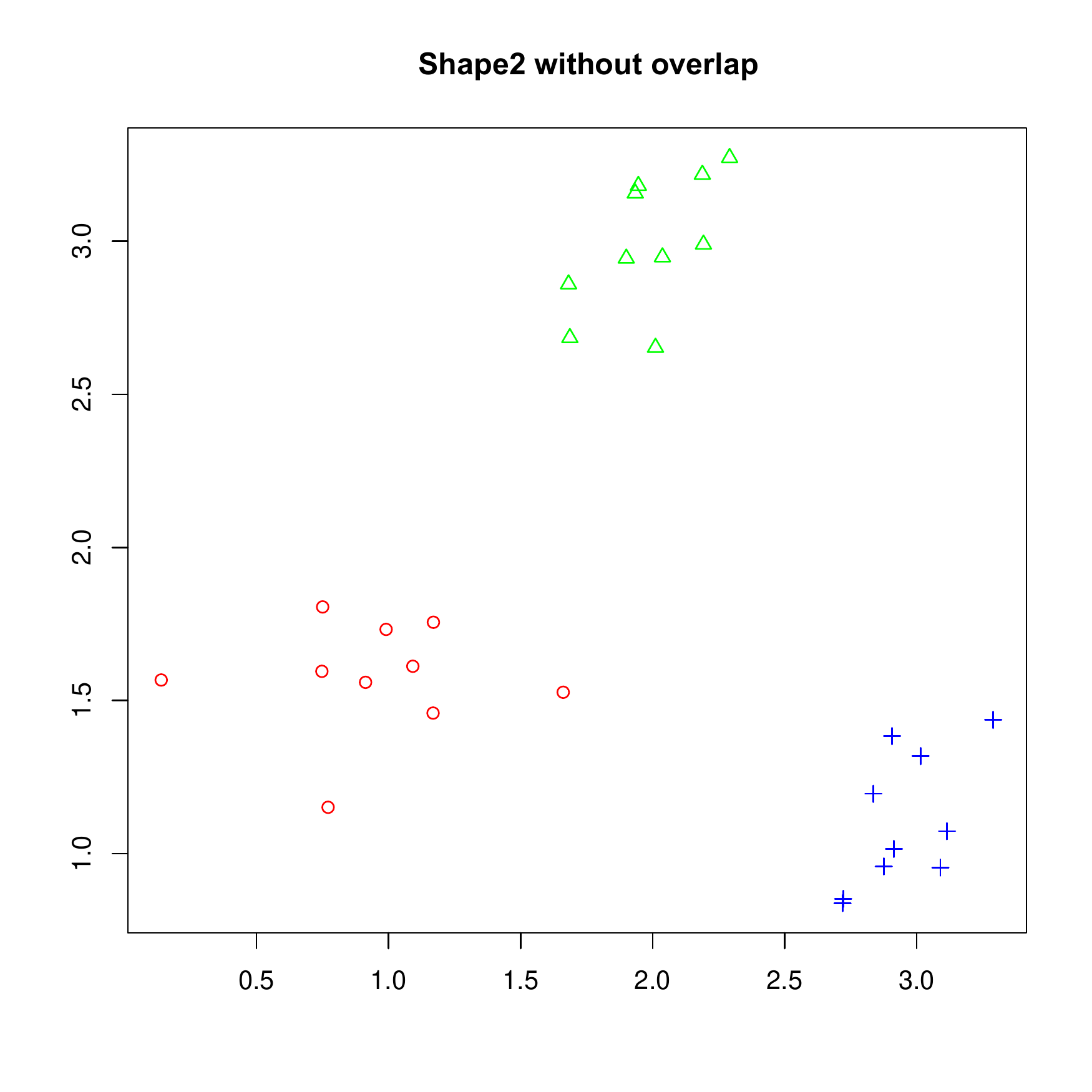}
\includegraphics[scale=0.3]{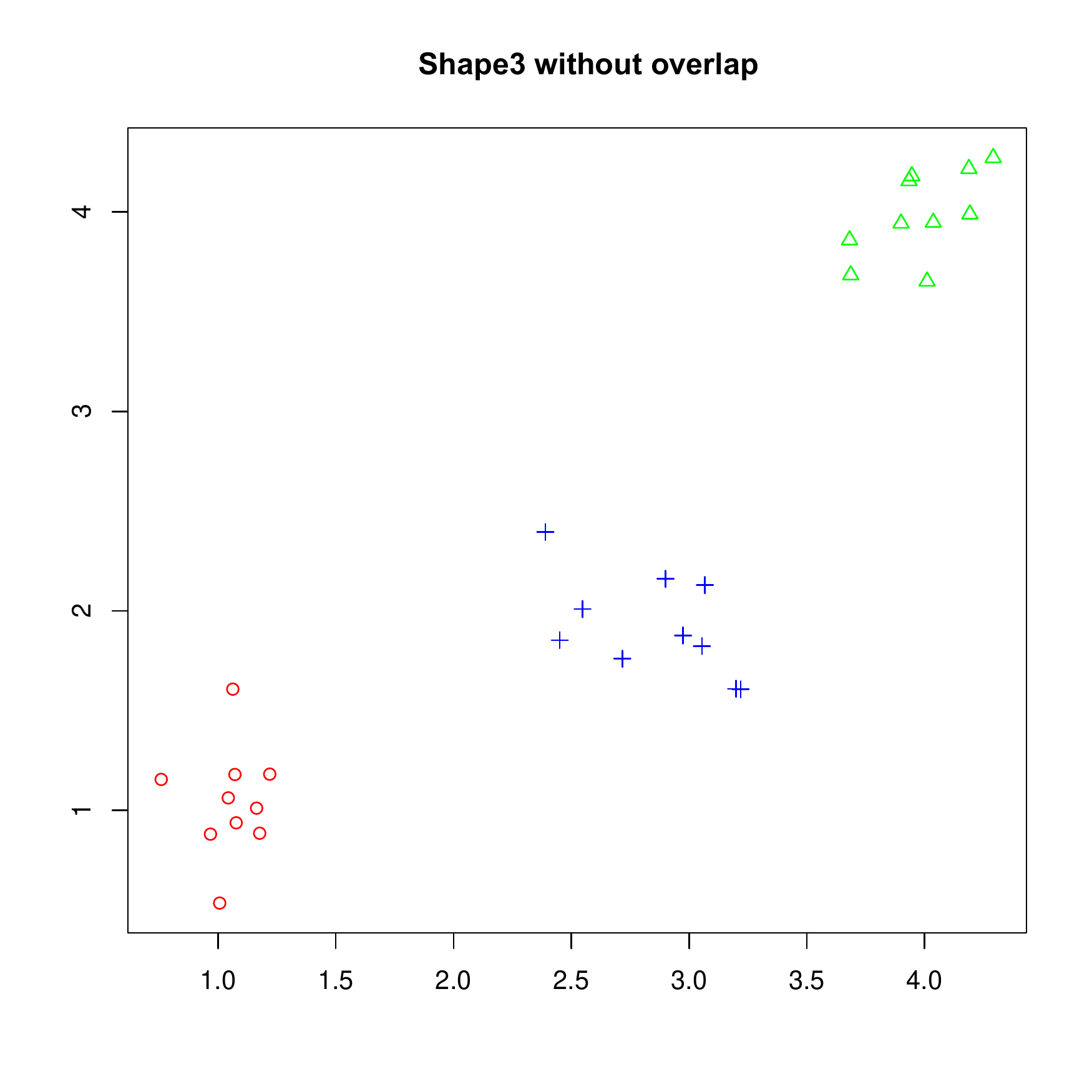}
\par\end{centering}
\centering{}
\caption{Vectorial data examples, each with 3 clusters represented by different point symbols of different colors}
\label{type and ovl}
\end{figure*}

\begin{table*}
\begin{centering}
\begin{tabular}{|l|l|l|}
\hline
\textbf{(shape, overlap)} & \textbf{Mean} & \textbf{Variance-Covariance (}$\Sigma_1$, $\Sigma_2$, $\Sigma_3$\textbf{)}\tabularnewline
\hline
\multirow{3}{*}{(1, with)} & $\mu_1=(1.1,1.1)^T$ & \multirow{3}{*}{$\begin{bmatrix}0.1&-0.03\\-0.03&0.1\end{bmatrix},\begin{bmatrix}0.15&-0.09\\-0.09&0.15\end{bmatrix},\begin{bmatrix}0.15&-0.09\\-0.09&0.15\end{bmatrix}$}\tabularnewline
 & $\mu_2=(2.1,2.3)^T$ & \tabularnewline
 & $\mu_3=(3.3,1.1)^T$ & \tabularnewline
\hline
\multirow{3}{*}{(2, with)} & $\mu_1=(1.2,1.2)^T$ & \multirow{3}{*}{$\begin{bmatrix}0.2&-0.1\\-0.1&0.2\end{bmatrix},\begin{bmatrix}0.1&0.05\\0.05&0.1\end{bmatrix},\begin{bmatrix}0.1&0.05\\0.05&0.1\end{bmatrix}$}\tabularnewline
 & $\mu_2=(1.4,2.4)^T$ & \tabularnewline
 & $\mu_3=(2.4,1)^T$ & \tabularnewline
\hline
\multirow{3}{*}{(3, with)} & $\mu_1=(1,0.6)^T$ & \multirow{3}{*}{$\begin{bmatrix}0.2&0.05\\0.05&0.2\end{bmatrix},\begin{bmatrix}0.2&0.05\\0.05&0.2\end{bmatrix},\begin{bmatrix}0.25&-0.12\\-0.12&0.25\end{bmatrix}$}\tabularnewline
 & $\mu_2=(2.5,2.5)^T$ & \tabularnewline
 & $\mu_3=(2.25,1)^T$ & \tabularnewline
\hline
\multirow{3}{*}{(1, without)} & $\mu_1=(1.1,1.1)^T$ & \multirow{3}{*}{$\frac{1}{3}\begin{bmatrix}0.1&-0.02\\-0.02&0.1\end{bmatrix},\frac{1}{3}\begin{bmatrix}0.15&-0.03\\-0.03&0.15\end{bmatrix},\frac{1}{3}\begin{bmatrix}0.15&-0.03\\-0.03&0.15\end{bmatrix}$}\tabularnewline
 & $\mu_2=(2.1,2.5)^T$ & \tabularnewline
 & $\mu_3=(3.5,1.1)^T$ & \tabularnewline
\hline
\multirow{3}{*}{(2, without)} & $\mu_1=(1,1.5)^T$ & \multirow{3}{*}{$\frac{1}{3}\begin{bmatrix}0.2&-0.03\\-0.03&0.2\end{bmatrix},\frac{1}{3}\begin{bmatrix}0.1&0.02\\0.02&0.1\end{bmatrix},\frac{1}{3}\begin{bmatrix}0.1&0.02\\0.02&0.1\end{bmatrix}$}\tabularnewline
 & $\mu_2=(2,3)^T$ & \tabularnewline
 & $\mu_3=(3,1)^T$ & \tabularnewline
\hline
\multirow{3}{*}{(3, without)} & $\mu_1=(1,1)^T$ & \multirow{3}{*}{$\frac{1}{3}\begin{bmatrix}0.1&0.02\\0.02&0.1\end{bmatrix},\frac{1}{3}\begin{bmatrix}0.1&0.02\\0.02&0.1\end{bmatrix},\frac{1}{3}\begin{bmatrix}0.2&-0.03\\-0.03&0.2\end{bmatrix}$}\tabularnewline
 & $\mu_2=(4,4)^T$ & \tabularnewline
 & $\mu_3=(3,2)^T$ & \tabularnewline
\hline
\end{tabular}
\par\end{centering}
\centering{}
\caption{Parameters for generating the data in Fig. \ref{type and ovl}}
\label{para for type and ovl}
\end{table*}
\par\end{center}

For the network data $\boldsymbol Y$, the difficulty of clustering is controlled by the relative magnitude of the linking probabilities in $\boldsymbol \Psi$. As we can expect, clustering a network would be easier if there are more within-cluster edges and less between-cluster edges. Reflecting on the probability matrix $\boldsymbol \Psi$, the ``noise'' level depends on whether the diagonal elements are significantly larger than off-diagonal elements. We test on network examples with both high noise and low noise. Fig. \ref{high low noise} shows two examples. The corresponding probability matrices are provided in Table \ref{para low and high}.

\begin{center}
\begin{figure*}
\begin{centering}
\includegraphics[scale=0.4]{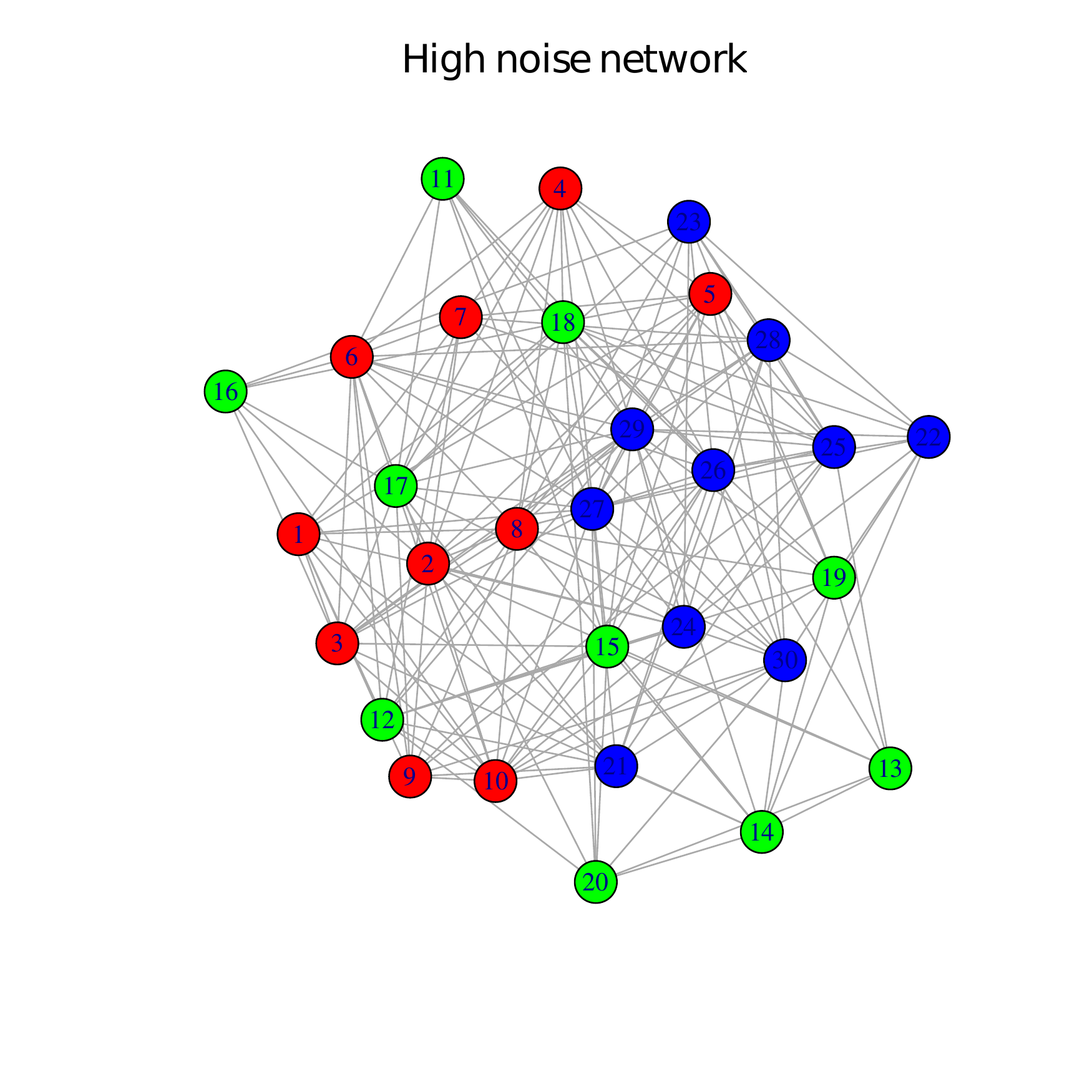}
\includegraphics[scale=0.4]{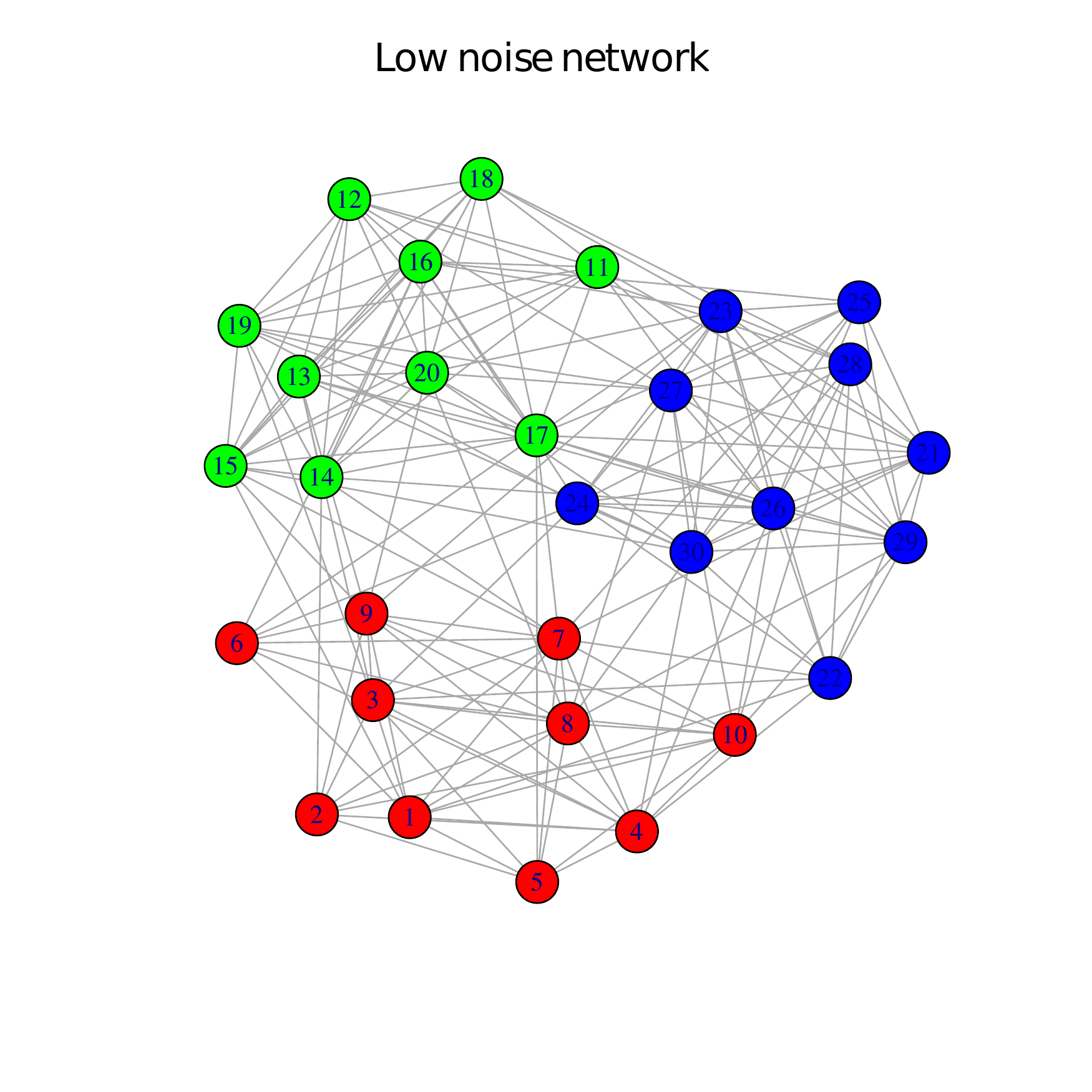}
\par\end{centering}
\centering{}
\caption{High noise (left) and low noise (right) networks}
\label{high low noise}
\end{figure*}

\begin{table}[H]
\begin{doublespace}
\noindent \begin{centering}
\begin{tabular}{|c|c|c|}
\hline
 & \textbf{High noise} & \textbf{Low noise}\tabularnewline
$\boldsymbol \Psi$ & $\begin{pmatrix}0.6&0.25&0.35\\0.25&0.65&0.35\\0.35&0.35&0.65\end{pmatrix}$ & $\begin{pmatrix}0.8&0.15&0.2\\0.15&0.9&0.25\\0.2&0.25&0.9\end{pmatrix}$\tabularnewline
\hline
\end{tabular}
\par\end{centering}
\noindent
\centering{}
\caption{Probability matrix of the networks in Fig. \ref{high low noise}}
\label{para low and high}\end{doublespace}
\end{table}
\par\end{center}

\subsection{Accuracy measure}
\label{accur} To evaluate our method and compare with other
methods using simulated data where the true cluster memberships
are known, we adopt the widely used Adjusted Rand Index (ARI)
\citet{hubert1985comparing} to measure the consistency between the
inferred clustering and the ground truth. For each pair of true
cluster label $\boldsymbol C_{true}$ and inferred $\boldsymbol
C_{inf}$, a contingency table is established first and ARI is then
calculated according to the formula in
\citet{hubert1985comparing}. An ARI with value 1 means the
clustering result is completely correct compared to the truth, and
ARI will be less than 1 or even negative when objects are wrongly
clustered. An advantage of using ARI is that we don't need to
explicitly match the labels of the two comparing clusterings. What
we are interested in is whether a certain set of objects belong to
a same cluster rather than the label number itself.

Since currently no similar methods perform simultaneous clustering for both vectorial and network data, we compare our method with results from individual data types and their intuitive combination. Methods in \citet{Rossi2006GMM} and \citet{schmidt2013nonparametric} are used to cluster the vectorial data (denote as ``Vec'') and network data (denote as ``Net''), respectively. These clustering results are stored in $\boldsymbol C_{vec}$ and $\boldsymbol C_{net}$. An intuitive way to combine them is the idea of multiple voting as used by ensemble methods \citep{Dietterich:2000:LectureNotes,Fred:Jain:2002:ICPR,Strehl:Ghosh:2003:JMLR}, which combine different clustering results in an post-processing fashion. In our scenario, we construct a contingency table between $\boldsymbol C_{vec}$ and $\boldsymbol C_{net}$ to find the best mapping, then calculate an average ARI as compared to the truth (denoted as ``Combine''). As an extra reference, we also take the better clustering between ``Net'' and ``Vec'' (denoted as ``Oracle'') as if we know which data type we shall trust. Thus, ``Oracle'' represents the upper bound of the performance for post-processing ensemble methods on our scenario, but it is not really achievable since we do not know which one to trust before we know the true clustering.

\subsection{Simulation results in cases with different data conditions}
\label{simu1}
We simulated data from nine cases listed in Table \ref{K=00003D3 N=00003D30}, which represents different combinations of the vectorial and network conditions. In this first experiment, we set $K=3$ and $N=30$ with each cluster containing 10 objects. For each of the nine cases, 10 independent data sets ($\boldsymbol X$ and $\boldsymbol Y$) are generated.

When running our Shared Clustering method, we observe that 1000
iterations seem sufficient for the MCMC algorithm to converge for
cases in this subsection, while more iterations are needed in
later experiments such as high dimensional cases. Thus we run our
MCMC algorithm for 2000 iterations and MAP is used to calculate an
ARI as the accuracy of the MCMC chain. For each of the cases in
Table \ref{K=00003D3 N=00003D30}, ten independent datasets are
generated. For each of the dataset, we independently run ten MCMC
chains and the median ARI of the ten chains is used as the
accuracy of the algorithm on this dataset. The mean and standard
deviation of the ten ARIs from the ten datasets are used to
represent the performance of the algorithm on a specific case. The
same datasets are used to assess the performance of ``Vec'',
``Net'', ``Combine'' and ``Oracle''. The simulation results are
summarized in Table \ref{K=00003D3 N=00003D30}.

\begin{center}
\begin{table*}
\begin{onehalfspace}
\noindent \begin{centering}
\begin{tabular}{|c|c|c|c|c|c|c|c|c|}
\hline
\textbf{Case} & \textbf{noise} & \textbf{overlap} & \textbf{shape} & \textbf{Shared} & \textbf{Combine} & \textbf{Oracle} & \textbf{Net} & \textbf{Vec}\tabularnewline
\hline
1 & low & with & 1 & \textbf{1(0)} & 0.742(0.124) & 1(0) & 1(0) & 0.580(0.166)\tabularnewline
\hline
2 & low & with & 2 & \textbf{1(0)} & 0.531(0.053) & 1(0) & 1(0) & 0.306(0.075)\tabularnewline
\hline
3 & low & with & 3 & \textbf{1(0)} & 0.607(0.055) & 1(0) & 1(0) & 0.418(0.085)\tabularnewline
\hline
4 & high & without & 1 & \textbf{0.956(0.080)} & 0.468(0.191) & 0.907(0.079) & 0.270(0.253) & 0.907(0.079)\tabularnewline
\hline
5 & high & without & 2 & \textbf{0.951(0.093)} & 0.501(0.194) & 0.955(0.055) & 0.294(0.242) & 0.955(0.055)\tabularnewline
\hline
6 & high & without & 3 & \textbf{0.971(0.065)} & 0.517(0.191) & 1(0) & 0.287(0.247) & 1(0)\tabularnewline
\hline
7 & high & with & 1 & \textbf{0.884(0.121)} & 0.346(0.211) & 0.588(0.189) & 0.286(0.251) & 0.575(0.190)\tabularnewline
\hline
8 & high & with & 2 & \textbf{0.672(0.213)} & 0.221(0.147) & 0.381(0.184) & 0.297(0.255) & 0.305(0.083)\tabularnewline
\hline
9 & high & with & 3 & \textbf{0.720(0.197)} & 0.272(0.133) & 0.480(0.115) & 0.291(0.217) & 0.418(0.085)\tabularnewline
\hline
\end{tabular}
\par\end{centering}
\noindent
\centering{}
\caption{Clustering performance on the nine cases with $K=3$ and $N=30$. The mean (sd) ARI of each case is calculated from 10 independent trials}
\label{K=00003D3 N=00003D30}
\end{onehalfspace}
\end{table*}
\par\end{center}

From Table \ref{K=00003D3 N=00003D30}, one can conclude that when one data type, ($\boldsymbol X$ or $\boldsymbol Y$) was ``clean'' (easy to cluster) while the other data type was ``dirty'' (as in Case 1-3 and Case 4-6), the single-data-type method corresponding to the clean data would give outstanding performance despite that the other single-data-type method was nearly of no use in terms of clustering. The ``Combine'' method was naturally deteriorated by the ``dirty'' side. However, Shared Clustering has the ability to take advantage of the clean data meanwhile largely prevent the negative effects from the dirty data. Its accuracy is similar to that of ``Oracle'' and much better than ``Combine''. When both two types of data were ``dirty'' (as in Case 7-9), neither of the single-data-type methods could work. But again, Shared Clustering performed significantly better than single-data-type methods, ``Combine'' and ``Oracle''.

\subsection{Simulation results in large number of objects}
\label{simu2} Observing the relatively low ARIs when both
$\boldsymbol X$ and $\boldsymbol Y$ were ``dirty'', we were
interested in whether a larger number of objects ($N$) could
increase clustering accuracy. Thus we further tested three cases
with $N=90$ and each cluster containing 30 objects. Simulation
settings for vectorial data remained unchanged as in Case 7-9,
however, the original ``high noise'' setting for network data was
no longer that ``noisy'' under the increased number of objects
since clustering would be surely easier with more connections.
Therefore, for cases with 30 objects in each cluster, instead of
using the ``high'' noise setting in Table \ref{para low and high},
we define $\boldsymbol
\Psi=\begin{pmatrix}0.55&0.3&0.4\\0.3&0.6&0.4\\0.4&0.4&0.6\end{pmatrix}$
as ``very high'' noise level, to roughly match the difficulty of
network data with the corresponding vectorial data. The results
are shown in Table \ref{K=00003D3 N=00003D90}. As expected, the
performance based on the network data alone is dramatically
increased although we increased the noise level, but the
performance based on the vectorial data alone is hardly changed.
The Shared Clustering also showed an improved performance with
higher ARIs and lower standard deviations.
\begin{center}
\begin{table*}
\begin{onehalfspace}
\noindent
\begin{centering}
\begin{tabular}{|c|c|c|c|c|c|c|c|c|}
\hline \textbf{Case} & \textbf{noise} & \textbf{overlap} &
\textbf{type} & \textbf{Shared} & \textbf{Combine} &
\textbf{Oracle} & \textbf{Net} & \textbf{Vec}\tabularnewline
\hline 10 & very high & with & 1 & \textbf{0.911(0.047) } &
0.410(0.107)  & 0.626(0.146)  & 0.626(0.146)  & 0.369(0.071)
\tabularnewline \hline 11 & very high & with & 2 &
\textbf{0.884(0.058) } & 0.391(0.062) & 0.630(0.145)  &
0.630(0.145)  & 0.304(0.034) \tabularnewline \hline 12 & very high
& with & 3 & \textbf{0.833(0.066) } & 0.465(0.058)  & 0.648(0.109)
& 0.624(0.156)  & 0.415(0.047) \tabularnewline \hline
\end{tabular}
\par\end{centering}
\noindent
\centering{}
\caption{Clustering performance on the three cases with $K=3$ and $N=90$. The mean (sd) ARI of each case is calculated from 10 independent trials}
\label{K=00003D3 N=00003D90}
\end{onehalfspace}
\end{table*}
\par\end{center}

\subsection{Simulation results in large number of clusters}
\label{simu3}
We were also interested in the performance of the method when $K$ increases. Hence we extended our experiments to test some cases with a larger cluster number $K=10$. More specifically, four of the ten mean vectors fall into the rectangular region located by point (1, 1), (1, 4), (4, 1) and (4, 4); three of the ten fall into the rectangular region located by (4, 7), (4, 10), (7, 7) and (7, 10); the other three are in the region located by (6, 3), (6, 8), (10, 3) and (10, 8). And the covariance matrices of the ten clusters are randomly assigned among three different types: non-correlated $\begin{pmatrix}0.5&0\\0&0.5\end{pmatrix}$, positively-correlated $\begin{pmatrix}0.5&0.4\\0.4&0.5\end{pmatrix}$, and negatively-correlated $\begin{pmatrix}0.5&-0.4\\-0.4&0.5\end{pmatrix}$. The motivation of these designs is to avoid cases where many clusters crushed together or many clusters separated too far, which are either impossible or too easy for clustering. Fig. \ref{Xplots K=00003D10} provides two plots of examples for vectorial data with $N=100$ and $N=300$ respectively.

\begin{onehalfspace}
\noindent \begin{center}
\begin{figure*}
\begin{onehalfspace}
\noindent
\begin{centering}
\includegraphics[scale=0.4]{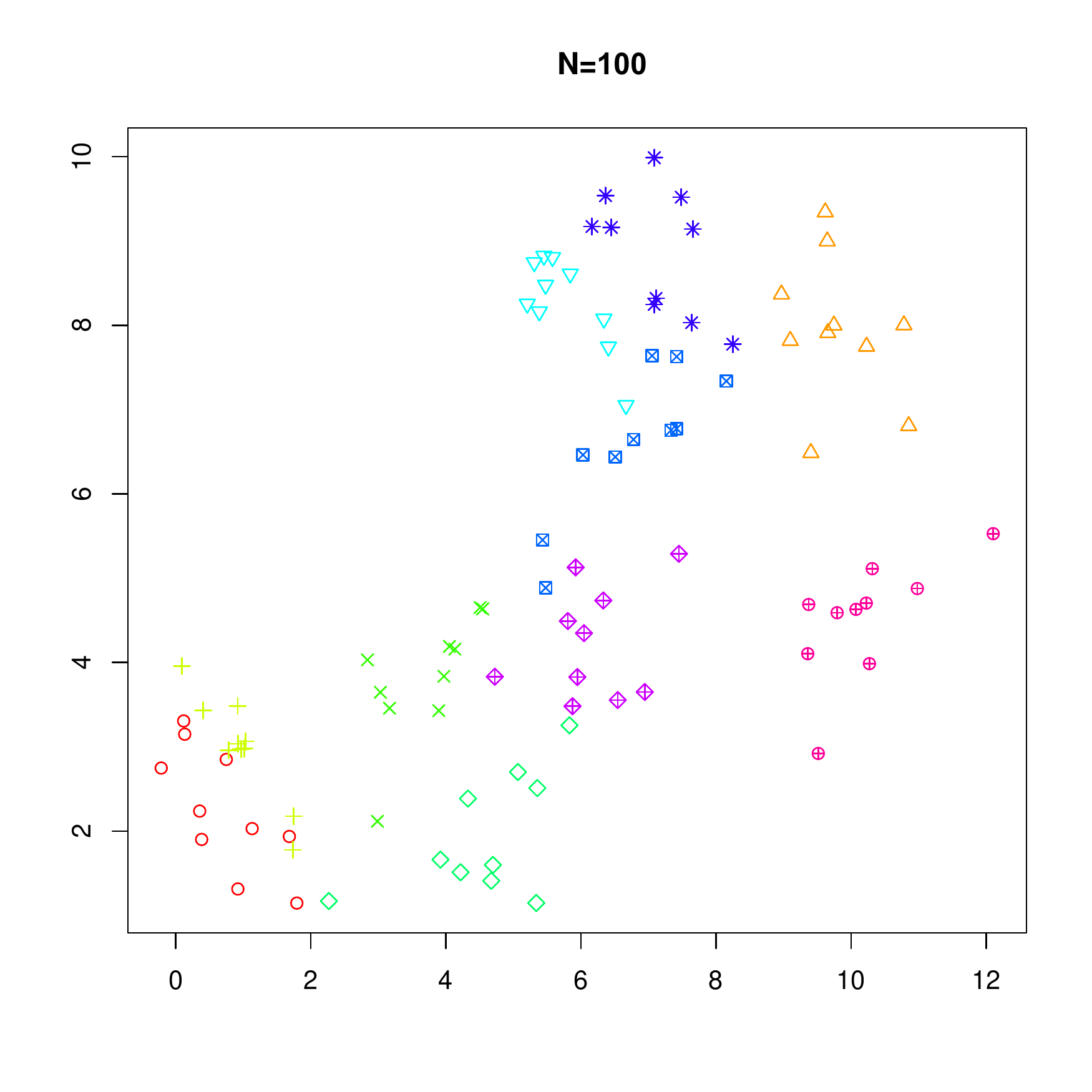}
\includegraphics[scale=0.4]{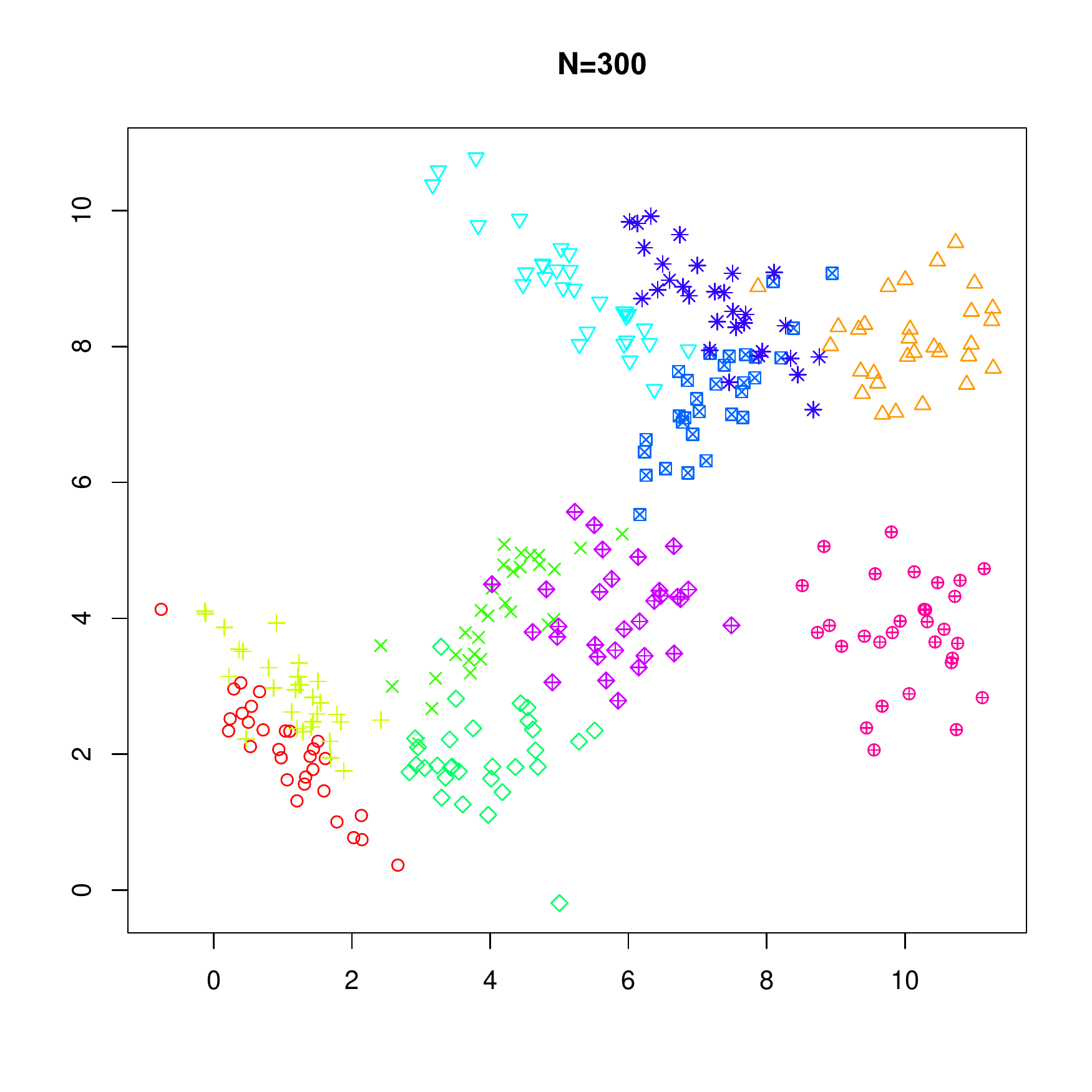}
\par\end{centering}
\noindent
\centering{}
\caption{Examples of $\boldsymbol X$ with $N=100$ and $N=300$, each with 10 clusters represented by different point symbols of different colors}
\label{Xplots K=00003D10}
\end{onehalfspace}
\end{figure*}
\par\end{center}
\end{onehalfspace}

For network data, we used newly defined levels of noise called ``moderate'' and ``hard.'' The ``moderate'' level is designed to be relatively easier for clustering than the ``hard'' level. The two 10-by-10 probability matrices are presented in Table S1 in the supplementary document.

We tested on four cases for this study on the effect of K, two with ten objects in each cluster ($N=100$) and the other two with thirty objects in each cluster ($N=300$). The tests are similar to those with three clusters, namely for each case we use the three methods and calculate the ARI means and standard deviations. The experiment results are shown in Table \ref{large clusters K=00003D10}.

\begin{onehalfspace}
\noindent
\begin{center}
\begin{table*}
\begin{onehalfspace}
\noindent
\begin{centering}
\begin{tabular}{|c|c|c|c|c|c|c|c|}
\hline
\textbf{Case} & \textbf{N} & \textbf{noise} & \textbf{Shared} & \textbf{Combine} & \textbf{Oracle} & \textbf{Net} & \textbf{Vec}\tabularnewline
\hline
13 & 100 & moderate & \textbf{0.805(0.058)} & 0.449(0.021) & 0.635(0.043) & 0.635(0.043) & 0.457(0.043)\tabularnewline
\hline
14 & 100 & messy & \textbf{0.496(0.073)} & 0.126(0.019) & 0.450(0.047) & 0.036(0.017) & 0.450(0.047)\tabularnewline
\hline
15 & 300 & moderate & \textbf{0.869(0.034)} & 0.532(0.050) & 0.798(0.037) & 0.798(0.037) & 0.481(0.059)\tabularnewline
\hline
16 & 300 & messy & \textbf{0.913(0.053)} & 0.327(0.055) & 0.496(0.057) & 0.340(0.135) & 0.479(0.059)\tabularnewline
\hline
\end{tabular}
\par\end{centering}
\noindent
\centering{}
\caption{Clustering performance on the four cases with $K=10$. The mean (sd) ARI of each case is calculated from 10 independent trials}
\label{large clusters K=00003D10}\end{onehalfspace}
\end{table*}
\par\end{center}
\end{onehalfspace}

The previous experiments on $K=3$ have shown that network clustering on a small number of objects performs badly when the ``noise'' is relatively high. When the number of clusters grows to ten, the situation became even worse. This can be clearly seen in Case 14, where the clustering on network alone completely failed and even the Shared Clustering method could not improve the accuracy since no extra information is provided by the network data. However, if we increased the number of objects (Case 16), Shared Clustering was again showing a big advantage.

\subsection{Simulation results in higher dimensional vectorial data}
\label{simu4} In the real world, vectorial data are more likely to
have more than two dimensions. Here we present two experiments for
higher dimensional $\boldsymbol X$, with dimension $q=5$ and $20$
respectively. Besides $\mu_k$ and $\Sigma_k$, all the other
parameter settings, including number of clusters, number of
objectives, and noise level of network data are identical to those
in Case 10-12, as we purely attempt to examine the effects of
higher dimensions. Numerical details of the mean vectors ($\mu_k$)
of vectorial data for $q=5$ and $q=20$ are shown in Table S2 and
S3 respectively in  the supplementary document. The mean values of
the first 5 dimensions in the $q=20$ case are set as the same as
the corresponding mean in the $q=5$ case. For the covariance
matrices, all diagonal elements are set as 1 and off-diagonal
elements are sampled uniformly between -0.05 and 0.05.

The experiments results are listed in Table \ref{high dim}. Again
for each case, we used 10 independent datasets to test our method.
The large dimensions required more MCMC iterations to get
converged samples. For $q=5$, each chain was run 3000 iterations
with the first 2000 as burn-in; for $q=20$, the numbers are 4000
with 3000 burn-in. From Table \ref{high dim}, we observed that as
the dimension grew from $q=5$ to $q=20$, with more supporting
clustering information from the extra dimensions, clustering for
vectorial data improved significantly and it made Shared
Clustering even better. In both cases, Shared Clustering embraced
significantly better ARI scores.

\noindent
\begin{center}
\begin{table*}
\begin{onehalfspace}
\noindent \begin{centering}
\begin{tabular}{|c|c|c|c|c|c|c|}
\hline
\textbf{Case} & \textbf{q} & \textbf{Shared} & \textbf{Combine} & \textbf{Oracle} & \textbf{Net} & \textbf{Vec}\tabularnewline
\hline
17 & 5 & \textbf{0.983(0.018)} & 0.723(0.044)  & 0.810(0.052)  & 0.698(0.098)  & 0.773(0.083) \tabularnewline
\hline
18 & 20 & \textbf{1(0)} & 0.627(0.246)  & 0.890(0.180)  & 0.705(0.100)  & 0.868(0.213) \tabularnewline
\hline
\end{tabular}
\par\end{centering}
\end{onehalfspace}
\caption{High dimensional experiments with $K=3$ and $N=90$ }
\label{high dim}
\end{table*}
\par\end{center}

\subsection{Selection of the number of clusters}
\label{selec} One advantage of our Bayesian approach is to
quantify the clustering uncertainty from the converged MCMC
sample. For each pair of objects $i$ and $j$, by counting the
times that they have a common cluster label among the sample, we
can estimate their pairwise co-clustering probability, which
indicates how likely the objects $i$ and $j$ are from the same
cluster. Repeating this counting process for all pairs of objects,
we get a $N$-by-$N$ pairwise co-clustering probability matrix.
This matrix is then processed to draw a heatmap, from which the
cluster structure can be easily visualized.

\begin{figure*}
    \begin{centering}
        \includegraphics[scale=0.3]{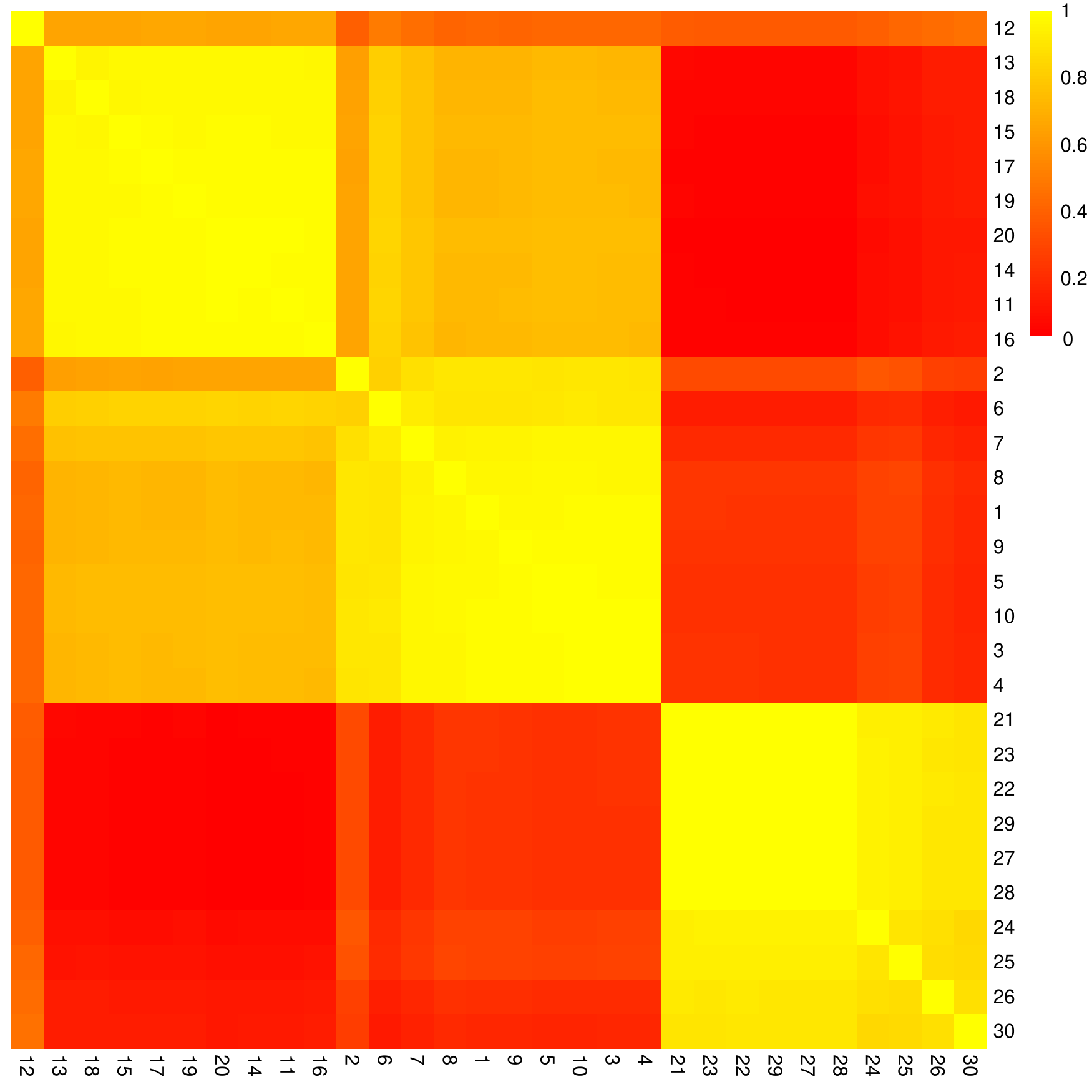}
        \includegraphics[scale=0.3]{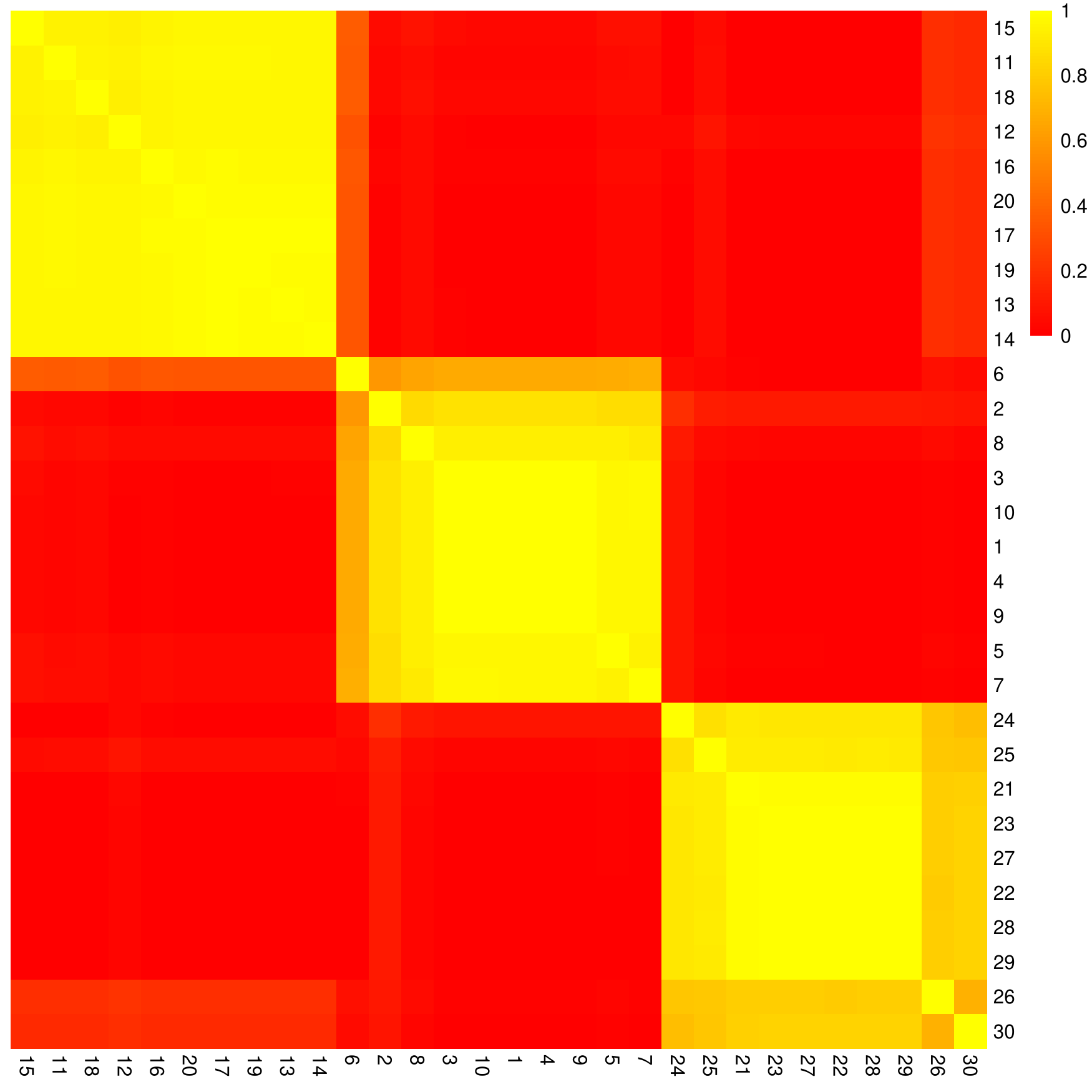}
        \includegraphics[scale=0.3]{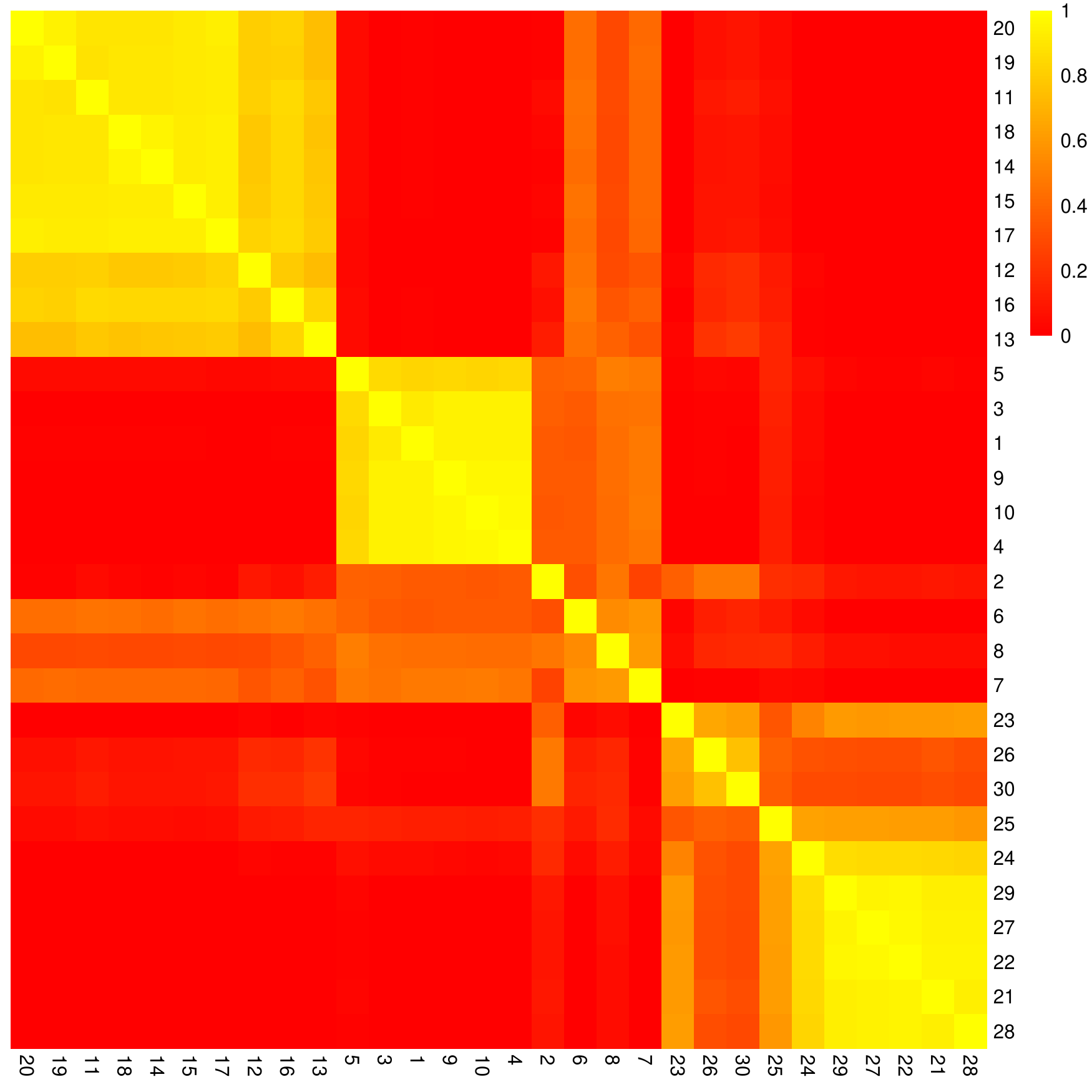}
        \par\end{centering}
    \centering{}
    \caption{Heatmaps from different numbers of clusters (from left to right: $K=2, 3, 4$), where $K=3$ is the true number}
    \label{heatmaps}
\end{figure*}

The above heatmap method can provide an intuitive way to select
the number of clusters. Take Case 7 in Section \ref{simu1} as an
example. We run our Shared Clustering with $K=2, 3, 4$ separately
until converge. Corresponding heatmaps are drawn in  Fig.
\ref{heatmaps} with the help of the R package ``pheatmap''
\citep{pheatmap}. From the heatmaps, one can draw a conclusion
that $K=3$ is the best choice in this case, because $K=3$ gives a
clearer cluster structure. When $K$ is set to be too big as in the
$K=4$ case, certain cluster in the $K-1$ heatmap is forced to
break, but which cluster to break is of uncertain, thus resulting
in blurry bars in the $K$ heatmap. When $K$ is set to be too small
as in the $K=2$ case, some clusters in the $K+1$ heatmap are
forced to merge, but which cluster to break is of uncertain, thus
resulting in a non-homogenous block in the $K$ heatmap.

\subsection{Prior sensitivity of the network parameters}
\label{sensi} We conducted sensitivity analysis for the prior
setting of the network parameter $\psi$ which follows Beta
distribution with shape parameters  $\beta_1$ and  $\beta_2$.
Instead of the prior setting mentioned in Section \ref{prior},
here we test the uniform prior with $\beta_1=\beta_2=1$. Case
10-12 in Section \ref{simu2} are re-done and the results are shown
in Table \ref{sensitivity analysis}. The performance indicates
that Shared Clustering behaved stably under different Beta
distribution priors.

\begin{center}
    \begin{table}
        \begin{onehalfspace}
            \noindent
            \begin{centering}
                \begin{tabular}{|c|c|c|}
                    \hline \textbf{Case} & Original & \textbf{$\beta_1=\beta_2=1$}\tabularnewline
                    \hline 10 & 0.911(0.047) & 0.967(0.032)
                    \tabularnewline \hline 11 & 0.884(0.058) & 0.877(0.058) \tabularnewline \hline 12 & 0.833(0.066) & 0.846(0.069) \tabularnewline \hline
                \end{tabular}
                \par\end{centering}
            \noindent
            \centering{}
            \caption{Clustering performance on case 10-12 with $\beta_1=\beta_2=1$. The mean (sd) ARI of each case is calculated from 10 independent trials}
            \label{sensitivity analysis}
        \end{onehalfspace}
    \end{table}
    \par\end{center}

\section{Real data experiment}
\label{reald} We tested our algorithm using a real gene dataset
used in \citet{gunnemann2010subspace}. The original processed data
in \citet{gunnemann2010subspace} contain 3548 genes with gene
interactions as edges; each gene has 115 gene expression values,
thus the dimension of the vectorial data is 115.
\citet{gunnemann2010subspace} used GAMEer to detect multiple
subnetworks from the whole large complex network. We aim at
checking whether the subnetworks from GAMer are clearly supported
by both the vectorial data and the network data. The selected
genes are listed in Table \ref{Real clusters}, and the gene IDs
are from \citet{gunnemann2010subspace}.

\noindent
\begin{center}
\begin{table*}
\begin{onehalfspace}
\noindent
\begin{centering}
\begin{tabular}{|c|ccccccccccc|}
\hline
Subnetwork 1 & 52 & 202 & 233 & 399 & 458 & 320 & 1078 & 1110 & 731 & 1345 & 1392\tabularnewline
 & 2096 & 1458 & 2432 & 2132 & 1384 & 3423 & 1702 &  &  &  & \tabularnewline
Subnetwork 2 & 352 & 337 & 391 & 398 & 410 & 460 & 485 & 411 & 1127 & 1213 & 1653\tabularnewline
Subnetwork 3 & 285 & 614 & 672 & 702 & 885 & 1117 & 2617 & 3382 & 3438 &  & \tabularnewline
\hline
\end{tabular}
\par\end{centering}
\noindent
\centering{}
\caption{Selected genes for clustering (gene ID identical to the data given by \citet{gunnemann2010subspace}) }
\label{Real clusters}
\end{onehalfspace}
\end{table*}
\par\end{center}

Due to the missing values contained in the original vectorial
data, all dimensions containing missing values for the 40 selected
genes are discarded, resulting in a 19 dimensional data set.
However, 19 is still larger than the number of genes in any of the
three subnetworks, which makes GMM under an ill condition
\citep{fraley2002model}. We thus employed PCA to reduce the
dimension of the vectorial data while maintaining most of the
variation in the data. The scree plot
\citep{mardia1979multivariate} of PCA is shown in Fig.
\ref{Screeplot}. According to the plot, we choose the first four
principal components as the finalized vectorial data.

\noindent
\begin{center}
\begin{figure}[H]
\noindent
\begin{centering}
\includegraphics[scale=0.5]{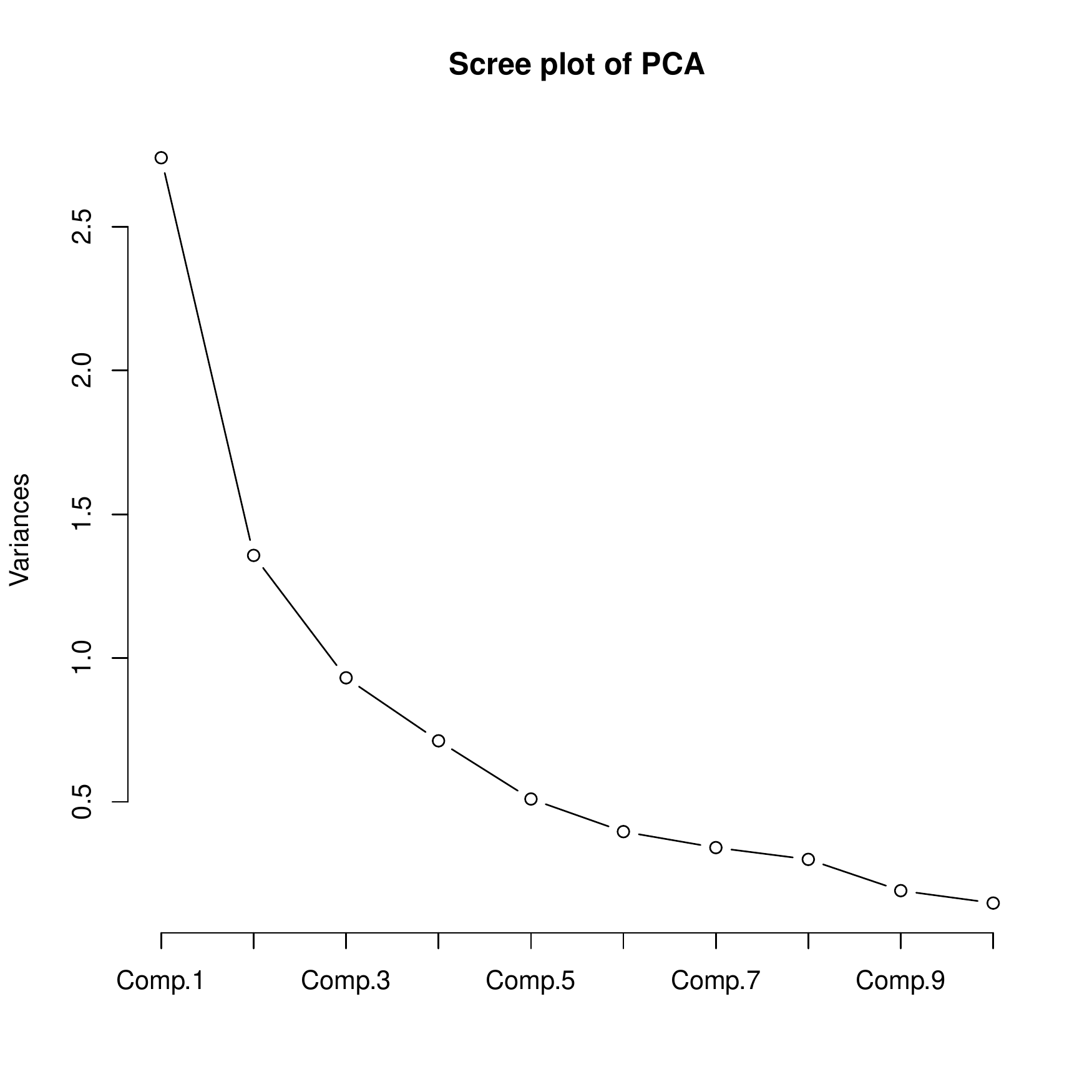}
\par\end{centering}
\noindent \centering{} \caption{Scree plot of the principal
components calculated from the 19 non-missing dimensions of the
selected 40 genes} \label{Screeplot}
\end{figure}

\par\end{center}

The interactions between genes are originally directed. We convert
the directed graph to undirected by simply considering any existed
link as an edge, namely $y_{ij}=1$ when there is an edge either
pointing from $i$ to $j$ or pointing from $j$ to $i$. The
processed undirected network is displayed in Fig. \ref{Real
network}.

\noindent
\begin{center}
\begin{figure}[H]
\noindent
\begin{centering}
\includegraphics[scale=0.5]{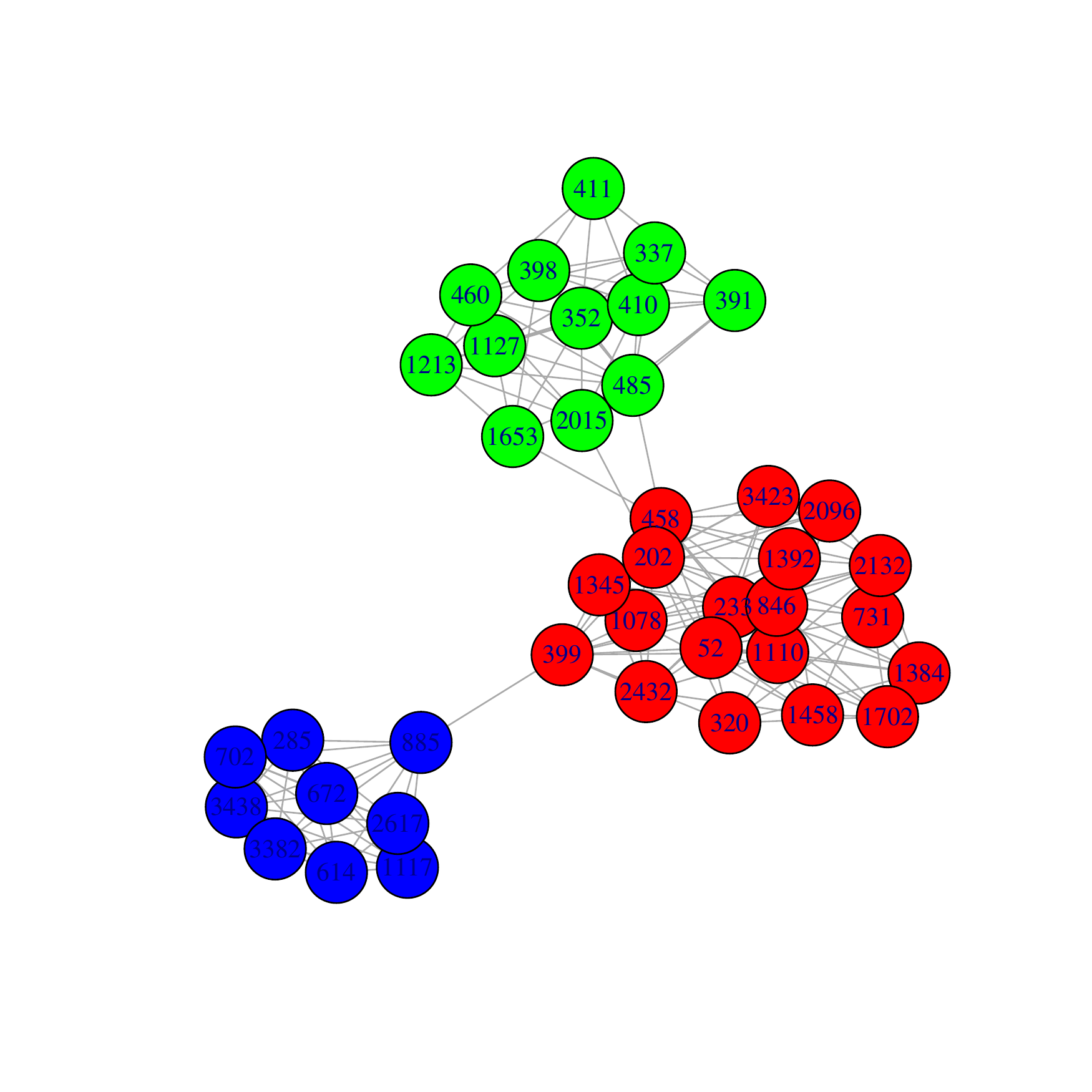}
\par\end{centering}
\noindent \centering{} \caption{Network of the selected genes for
clustering (gene ID identical to the data given by
\citet{gunnemann2010subspace})} \label{Real network}
\end{figure}
\par\end{center}

After running our algorithm with $K=2,3,4$, we used the heatmap
approach introduced in Section \ref{selec} to determine the most
reasonable $K$. The best number of clusters turned out to be three
which is consistent with \citet{gunnemann2010subspace}. Under
$K=3$, the clustering result from Shared Clustering fully
confirmed (ARI=1) the subnetwork memberships of the 40 genes
listed in Table \ref{Real clusters}.

In many real data problem, the dimension of the vectorial data is
bigger than the number of objects, which makes it difficult to fit
the GMM part of our model. In this example, we used PCA to reduce
the dimension while trying to maintain most variation of the data.
Other methods are also possible. For example, one can introduce a
shrinkage estimator or assume certain sparsity structure when
estimating the covariance matrix. One can also perform variable
selection when doing clustering \citep{Raftery2006selection}.

\section{Summary and discussion}
\label{summa} In this paper, we introduced the new probabilistic
integrative clustering method which can cluster vectorial and
relational data simultaneously. We introduced the Shared
Clustering model within a general framework and also provided a
specific Normal-Bernoulli model. A Gibbs sampling algorithm is
provided to perform the Bayesian inference. We ran intensive
simulation experiments to test the performance of Shared
Clustering by controlling various factors such as cluster size,
number of clusters, noise level of network data, shape and
dimension of vectorial data, etc. At the same time, a model
selection approach is discussed by using the MCMC sample. Finally
a gene subnetwork data set was employed to demonstrate the
applicability of the method in real world. The new joint
probabilistic model is characterized by a more efficient
information utilization, thus shows a better clustering
performance.

Although we mainly concerned undirected graphs for the network
data in this paper, SBM can handle directed graph by simply
loosing the symmetric requirement for the adjacency matrix
$\boldsymbol Y$ and the probability matrix $\boldsymbol \Psi$. In
this case, the edge variable $\mathbf{y}_{ij}$ and
$\mathbf{y}_{ji}$ are modelled as independent, and both upper and
lower triangle of $\boldsymbol Y$ are useful. The number of
parameters in $\boldsymbol \Psi$ increases from $K(K+1)/2$ to
$K^2$. Moreover, the edge variables can be extended beyond binary
ones. For instance, $\mathbf{y}_{ij}$ can be Poisson variables
when modeling count-weighted graphs as in
\citet{mariadassou2010uncovering}, or Normal variables if the
network data is continuous data.

Similar logic can be applied to the vectorial data part. The
vectorial data can be continuous, discrete or even mixed type. For
continuous and discrete types, the distribution assumption can be
chosen accordingly. As for mixed type, for instance, the vector
can be $\mathbf{x}_{i}=(\mathbf{x}_{i1},\mathbf{x}_{i2})^T$ where
$\mathbf{x}_{i1}$ is a vector with continuous data and
$\mathbf{x}_{i2}$ is a vector with discrete data. Then the
distribution of vectorial data $f(\cdot)$ mentioned in Section
\ref{probl} is the joint distribution of the two random vectors.
If $\mathbf{x}_{i1}$ and $\mathbf{x}_{i2}$ are independent,
$\mathbf{x}_{i1}\sim f_1(\cdot)$ and $\mathbf{x}_{i2}\sim
f_2(\cdot)$, we will have $f(\cdot)=f_1(\cdot)f_2(\cdot)$. In
summary, as stated in Equation \ref{probl}, depending on the
observed types of data available, Shared Clustering may handle
different combinations of $f(\cdot)$ and $g(\cdot)$.

Since this paper mainly studied one specific model, the
Normal-Bernoulli model, performance of Shared Clustering under
other distributions for network and vectorial data as mentioned
above is still waiting for examination. Besides the method of
selecting the number of clusters $K$ that we discussed in Section
\ref{selec}, future studies are also needed to conduct model
selection in a more principled way. For example $K$ may be treated
as a random variable and sampled by the Reversible-Jump MCMC
\citep{green1995reversible}. \citet{mcdaid2013improved} and \cite{friel2013bayesian} have
proposed faster techniques to tackle this issue avoiding the
computationally expensive Reversible-Jump MCMC. Also, in some
situations, user may need to solve the label switching problem
\citep{jasra2005markov} in the MCMC sample. The technique
developed in \citet{li2014pivotal} can be used to tackle this.

In our current model, we independently model vectorial data and
network data given the cluster labels, thus there is no trade-off
between $\boldsymbol X$ and $\boldsymbol Y$. However, under
certain circumstance, if one has subjective knowledge of how the
two parts should be weighted, a tuning parameter can be introduced
to control the contribution of the two types of data.
Specifically, let $\eta$ be the tuning parameter, the joint
posterior in Equation \eqref{full_post} can be re-written as a
weighted one:
\begin{equation}\begin{split}p(&\boldsymbol
P,\boldsymbol C,\boldsymbol \Phi,\boldsymbol \Psi|\boldsymbol
X,\boldsymbol Y, \eta)\\&\propto p(\boldsymbol X, \boldsymbol
\Phi|\boldsymbol C)^{\eta}p(\boldsymbol Y, \boldsymbol
\Psi|\boldsymbol C)^{1-\eta}p(\boldsymbol C, \boldsymbol
P).\end{split}\label{weighted_likelihood}\end{equation} Note that
$\eta$ is a pre-specified tuning parameter, not to be treated as
random variable in the Bayesian inference. Further studies is also
needed in this kind of extension of the Shared Clustering model.

In our model, we assume that both vectorial data and network data
share the same clustering labels, which is the base of performing
joint clustering. In reality, we may not know whether we can
assume this same clustering. Strict testing of this assumption is
still an open problem. In practice, we can compare the two
clustering produced by individual data type using ARI. If the ARI
is too small, we shall doubt the assumption and avoid jointly
modeling the two data sets.

\begin{acknowledgements}
This research is partially supported by two grants
from the Research Grants Council of the Hong Kong SAR (Project No. CUHK 400913 and 14203915). The R codes and supplementary documents in this paper are available at \\ \url{https://github.com/yunchuankong/SharedClustering}.

\end{acknowledgements}

\appendix
\section{Calculation of posterior distributions}
\label{app_poste}

For $k=1,\ldots,K$, the posteriors of $\phi_k=(\Sigma_k, \mu_k)$
are given by \begin{equation}\Sigma_k|\boldsymbol C,\boldsymbol
X\sim\mathrm{InverseWishart_q}(T+\tilde
S_{k},v_0+N_k),\label{app_sigma_post}\end{equation}\begin{equation}\mu_k|\Sigma_k,\boldsymbol
C,\boldsymbol X\sim
\mathrm{N}(\tilde\mu_k,(\alpha+N_k)^{-1}\Sigma_k),
\label{app_mu_post}\end{equation} where $N_k$ is the current
number of objects in cluster $k$ (we use ``current'' since
parameters are updated iteratively by Gibbs sampler), and if we
denote $\mathbf{\bar x_k}$ as the sample mean in this cluster and
$S_{k}$ as the corresponding Sum of Square Cross Products (SSCP)
matrix, $\tilde S_{k}=S_k+{\alpha N_k\over
\alpha+N_k}(\mathbf{\bar x_k}-\mu_0)(\mathbf{\bar x_k}-\mu_0)^T$
and $\tilde\mu_k={\alpha\mu_0+N_k\mathbf{\bar x_k}\over
\alpha+N_k}$ are the updating parameters in Gibbs sampling for GMM
\citep{murphy2012machine,Rossi2006GMM}. All the notations not
explained here are consistent with those in Section \ref{metho}
(the same below).

For $k_1,k_2=1,\ldots,K$, the posterior distribution of each individual $\psi_{k_1,k_2}$ is again Beta distribution. Denoting the number of $y_{ij}$ in the block under $\psi_{k_1,k_2}$ by $N_b$, the posterior is thus \begin{equation}\psi_{k_1,k_2}|\boldsymbol C,\boldsymbol Y\sim\mathrm{Beta}(\beta_1+\sum y_{ij},\beta_2+N_b-\sum y_{ij}),\label{app_psi_post}\end{equation}where $i,j=1,\ldots,N$.

Calculating the posterior of cluster label $\boldsymbol C$ needs
more consideration. Since in network data, the distribution of
individual edge variable $y_{ij}$ is influenced by cluster labels
of both the $i$-th object and the $j$-th object, the labels in
$\boldsymbol C$ are not mutually independent in the posterior.
Hence we need to sample each cluster label $c_i$ conditional on
``other labels'' (denoted as $C_{-i}$). Applying Bayes' rule, the
posterior distribution of $c_i$, $i=1,\ldots,N$ can be derived as
\begin{equation}\begin{split}p(& c_i=k|C_{-i},\boldsymbol
\Phi,\boldsymbol \Psi,\mathbf x_i,\mathbf y_i,\boldsymbol P)\\
&\propto p(\mathbf x_i,\mathbf y_i|\boldsymbol \Phi,\boldsymbol
\Psi,c_i=k,C_{-i})p(c_i=k|\boldsymbol P),
\end{split}\label{app_c_post}\end{equation} where $\mathbf
y_i=(y_{i1},\ldots,y_{i,i-1},y_{i,i+1},\ldots,y_{iN})$ represents
the set of all edge variables associated with object $i$. The
first term of the right hand side in Equation \eqref{app_c_post}
is the joint likelihood of $\mathbf x_i$ and $\mathbf y_i$ given
$c_i=k$. And the second term is just $p_k$.

The posterior of $\boldsymbol P$ is updated according to
$p(\boldsymbol P|\boldsymbol C)\propto p(\boldsymbol C|\boldsymbol
P)p(\boldsymbol P)$. Let $\tilde
a_k=a_k+\sum\limits_{i=1}^N\mathbbm{1}\{c_i=k\}$, $k=1,\ldots,K$,
then \begin{equation}\boldsymbol
P\sim\mathrm{Dirichlet}(\boldsymbol{\tilde
a})\label{app_p_post}\end{equation} is the posterior distribution
of $\boldsymbol P$.

\bibliographystyle{spbasic}
\bibliography{Ref}
\end{document}